\UseRawInputEncoding
\pdfoutput=1
\documentclass[10pt,final,a4paper,oneside,onecolumn]{article}

\edef\reptitle{White paper on\\LiDAR performance against selected Automotive Paints}
\edef\shorttitle{LiDAR performance against selected Automotive Paints}
\def\repauthor{James Lee Wei Shung\\Paul Hibbard\\Roshan Vijay\\Lincoln Ang Hon Kin\\Niels de Boer}

\makeatletter
\def\@maketitle{%
	\newpage
	\null
	\vspace{-8em}
	\begin{center}%
		\begin{tabular}{l c r}
			\multicolumn{3}{c}{\includegraphics[width=0.6\linewidth]{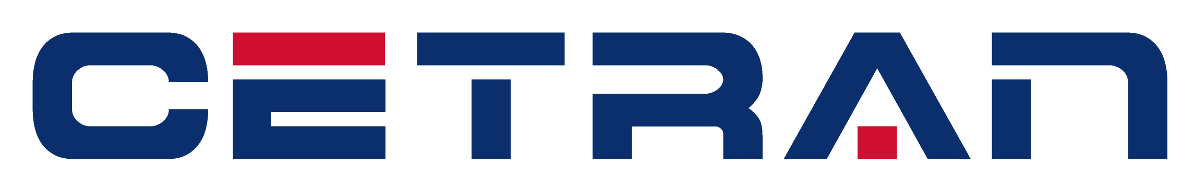}}\vspace{3em} \\
			\includegraphics[width=0.25\linewidth]{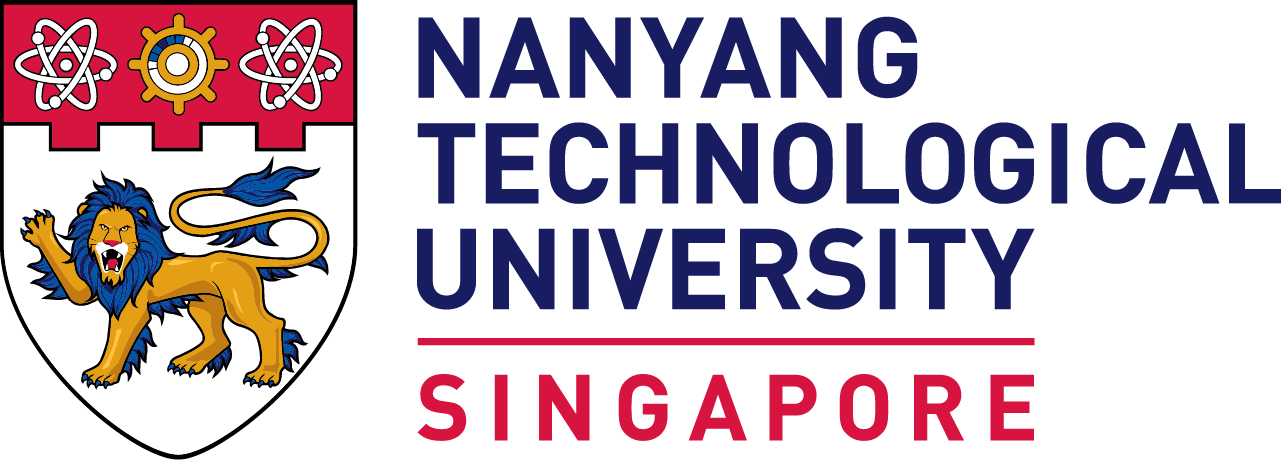} & \hspace{4em} &
			\includegraphics[width=0.3\linewidth]{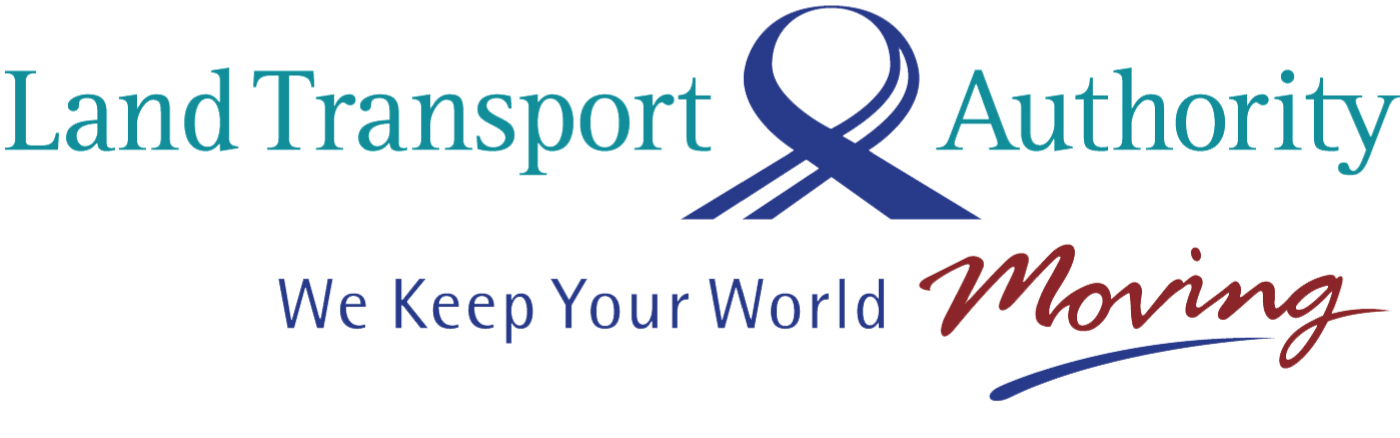}
		\end{tabular}\\
		\vskip 12em
		\let \footnote \thanks
		{\LARGE \@title \par}%
		\vskip 1em%
		{\large Version 1.0 from September 4, 2023}%
		\vskip 1.5em%
		{\large
			\lineskip .5em%
			\begin{tabular}[t]{c}%
				James Lee Wei Shung\\
				Paul Hibbard\\
				Roshan Vijay\\
				Lincoln Ang Hon Kin\\
				Niels de Boer
			\end{tabular}\par}%
	\end{center}%
	\par
	\vskip 1.5em}
\makeatother

\usepackage[utf8]{inputenc}
\usepackage[a4paper,left=2.0cm,right=2.0cm,top=3cm,bottom=3cm]{geometry} 
\usepackage[T1]{fontenc}                        
\usepackage{lmodern}                            
\usepackage{varwidth}
\usepackage[cmex10]{amsmath}                    
\usepackage{amssymb}                            
\usepackage{amsthm}                             
\usepackage{gensymb}
\usepackage{textcomp}
\usepackage{dsfont}                             
\usepackage{mathrsfs}                           
\usepackage[pdftex]{graphicx}                   
\usepackage{array}                              
\usepackage{upgreek}                            
\usepackage[style=ieee,doi=false,isbn=false,url=false,date=year,backend=biber,mincitenames=2,maxcitenames=2,mincitenames=1,uniquelist=false,uniquename=false,giveninits=true,sorting=nty]{biblatex}
\usepackage{stfloats}                           
\usepackage{float}
\usepackage{multirow}                           
\usepackage{subfig}                             
\usepackage{fancyhdr}                           
\usepackage[official,left]{eurosym}             
\usepackage{appendix}                           
\usepackage{xspace}                             
\usepackage{microtype}							
\usepackage{verbatim}                           
\usepackage[dutch,USenglish]{babel}             
\usepackage{wrapfig}                            
\usepackage{longtable}                          
\usepackage{pgfplots}                           
\usepackage{calc}                               
\usepackage{lscape}								
\usepackage[breaklinks=true,hidelinks,          
            bookmarksnumbered=true]{hyperref}   
\usepackage{vhistory}
\usepackage{makecell}
\usepackage[toc]{glossaries}                    
\usepackage[normalem]{ulem}                     
\usepackage{booktabs}							
\usepackage[bottom]{footmisc}                   
\usepackage{csvsimple}  						
\usepackage{makecell}  						
\usepackage{longtable}
\usepackage{acronym}                            
\usepackage{titlesec}
\usepackage{setspace}
\usepackage{changepage}
\usepackage{xcolor}

\usepackage{enumitem}

\usepackage{refcount}

\usepackage{rotating} 

\fancypagestyle{plain}{\fancyhf{}

	\setlength{\headheight}{17pt}
	\addtolength{\footskip}{-5pt}}

\newlength\figurewidth
\newlength\figureheight
\pgfplotsset{every axis/.append style={
		scaled y ticks=false,
		scaled x ticks=false,
		y tick label style={/pgf/number format/fixed},
		x tick label style={/pgf/number format/fixed},
		legend style={font=\small}},
	compat=1.9}                                 
\usetikzlibrary{arrows.meta}                    
\usetikzlibrary{external}                       

\theoremstyle{plain}    
\theoremstyle{definition}     
\theoremstyle{remark}\newtheorem{remarkenv}{Remark}[section]        
                       {\hfill$\lozenge$\end{remarkenv}}            

\graphicspath{{figures/}} 
\fnbelowfloat                                       
\hypersetup{pdftitle={\reptitle},
            pdfauthor={\repauthor},
            colorlinks=true,
            urlcolor=blue,
        	linkcolor=black,
        	citecolor=black}	               

\newlength\ndist                                
\newlength\nheight                              
\newlength\nwidth                                   
\newlength\nsep                                     

\newcommand\disclaimertext{%
	\begin{center} \textbf{Disclaimer} \end{center} 
	\footnotesize
        \begin{center} This white paper was developed with support from the Urban Mobility Grand Challenge Fund\\by the Land Transport Authority of Singapore (No. UMGC-L010). \end{center}
}
\newcommand\disclaimernotice{%
	\begin{tikzpicture}[remember picture,overlay]
	\node[anchor=south,yshift=100pt] at (current page.south) {\fbox{\parbox{\dimexpr\textwidth-\fboxsep-\fboxrule\relax}{\disclaimertext}}};
	\end{tikzpicture}%
}

\bibliography{references}

\begin{document}

\title{\Huge\textbf{\reptitle}}
\author{}

\maketitle

\vfill

\clearpage

\setlength\ndist{5em}   
\setlength\nheight{8em} 
\selectlanguage{USenglish}
\pagenumbering{roman}

\vfill

\clearpage

\tableofcontents
\disclaimernotice

\cleardoublepage

\pagenumbering{arabic}

\setcounter{table}{0}

\section{Abstract/Executive summary}
LiDAR (Light Detection and Ranging) is a useful sensing technique and an important source of data for autonomous vehicles (AVs). In this publication we present the results of a study undertaken to understand the impact of automotive paint on LiDAR performance along with a methodology used to conduct this study. Our approach consists of evaluating the average reflected intensity output by different LiDAR sensor models when tested with different types of automotive paints. The paints were chosen to represent common paints found on vehicles in Singapore.

The experiments were conducted with LiDAR sensors commonly used by autonomous vehicle (AV) developers and OEMs. The paints used were also selected based on those observed in real-world conditions. This stems from a desire to model real-world performance of actual sensing systems when exposed to the physical world. The goal is then to inform regulators of AVs in Singapore of the impact of automotive paint on LiDAR performance, so that they can determine testing standards and specifications which will better reflect real-world performance and also better assess the adequacy of LiDAR systems installed for local AV operations.

The tests were conducted for a combination of 13 different paint panels and 3 LiDAR sensors. In general, it was observed that darker coloured paints have lower reflection intensity whereas lighter coloured paints exhibited higher intensity values.

\section{Introduction}
\label{sec:introduction}
In recent years, development of robotic vehicles equipped with a dedicated Automated Driving System (ADS) \cite{standard2018j3016}, commonly known as autonomous vehicles (AVs), has received major attention from both the academic research community and mobility industry across the world. Automated Driving Systems on AVs perform the task of monitoring the driving environment in order to perform the dynamic driving task. The onboard sensing and perception (S\&P) module of an AV is therefore a crucial component which must function reliably and with acceptable quality and range of detection in order to ensure system safety. AVs typically rely on sensors such as LiDAR (Light Detection and Ranging), visible light cameras, RADAR (Radio Detection and Ranging) and ultrasound in order to sense and perceive the world around them. The data from these sensors is typically fused together to generate a comprehensive \textit{object-list} of static and dynamic obstacles around the vehicle. The fusion of sensor data ensures that the ADS has good spatial awareness with all-round coverage.

There have been numerous studies detailing the behaviour of camera systems under varied weather and environmental conditions as well as their response towards different materials, surfaces and textures. Moreover, most camera systems used for AV applications function in the visible light spectrum and have a similar range of colour vision as the human eye. Automotive LiDARs, on the other hand, typically operate in the infrared (IR) spectrum and have not been scrutinized as widely against automotive paint or other such materials and textures.

Much like humans `see' the world around them to make driving decisions, AVs use sensors such as LiDAR to `see' obstacles and vehicles around it. Vehicles may have different paints and coatings applied to them (either as a matter of personal preference or as legally mandated such as in the case of school buses in the United States \cite{NHTSA_Schoolbus}). All these paints and coatings are more or less highly detectable and visible to the human eye and by extension, visible spectrum cameras. However, their visibility from IR-spectrum LiDARs is an area where there have been relatively fewer number of studies and articles of literature. This white paper will explore automotive LiDAR sensors and its performance against different automotive paint coatings.

\subsection{Automotive LiDARs}
The sensors for automated driving can be split into two groups: active sensors and passive sensors. LiDAR and radar sensors are classified as active sensors as they emit the signals they intend to receive back as reflections, whereas, cameras are generally passive sensors, receiving signals that originate from independent sources ie. natural or artificial light reflected from objects \cite{Beiker2018}. LiDARs stand out because it is integral to most AV perception systems and compensates for the weaknesses of camera and radar sensors. It complements camera-based perception algorithms by providing detailed range measurements which may be limited otherwise, even with stereo cameras and may not be possible at all with a single camera. Many commercially available and open-source Automated Driving (AD) stacks such as Baidu Apollo \cite{Apollo_perception} or Autoware \cite{raju2019_autoware_apollo}, use LiDAR sensor data to generate a real-time object map of other road users and obstacles around the AV.

\subsubsection{Working principle}
LiDAR sensors \cite{li2020lidar} illuminate the surroundings by emitting laser light beams from an emitter, typically in the infrared (IR) spectrum of 903-1550nm. This region of the electromagnetic spectrum is not visible to the human eye \cite{li2020lidar}. The LiDAR sensor then uses a detector to detect reflected laser beams, from which it generates a point-cloud consisting of time-of-flight (ToF) calculated distances and reflected intensities of all objects around it \cite{behroozpour2017lidar}. A point-cloud is a set of points in 3D space which represents the real world as seen by the LiDAR measurements.

\begin{figure}[htb]
    \centering
    \centerline{
        \includegraphics[scale=0.50]{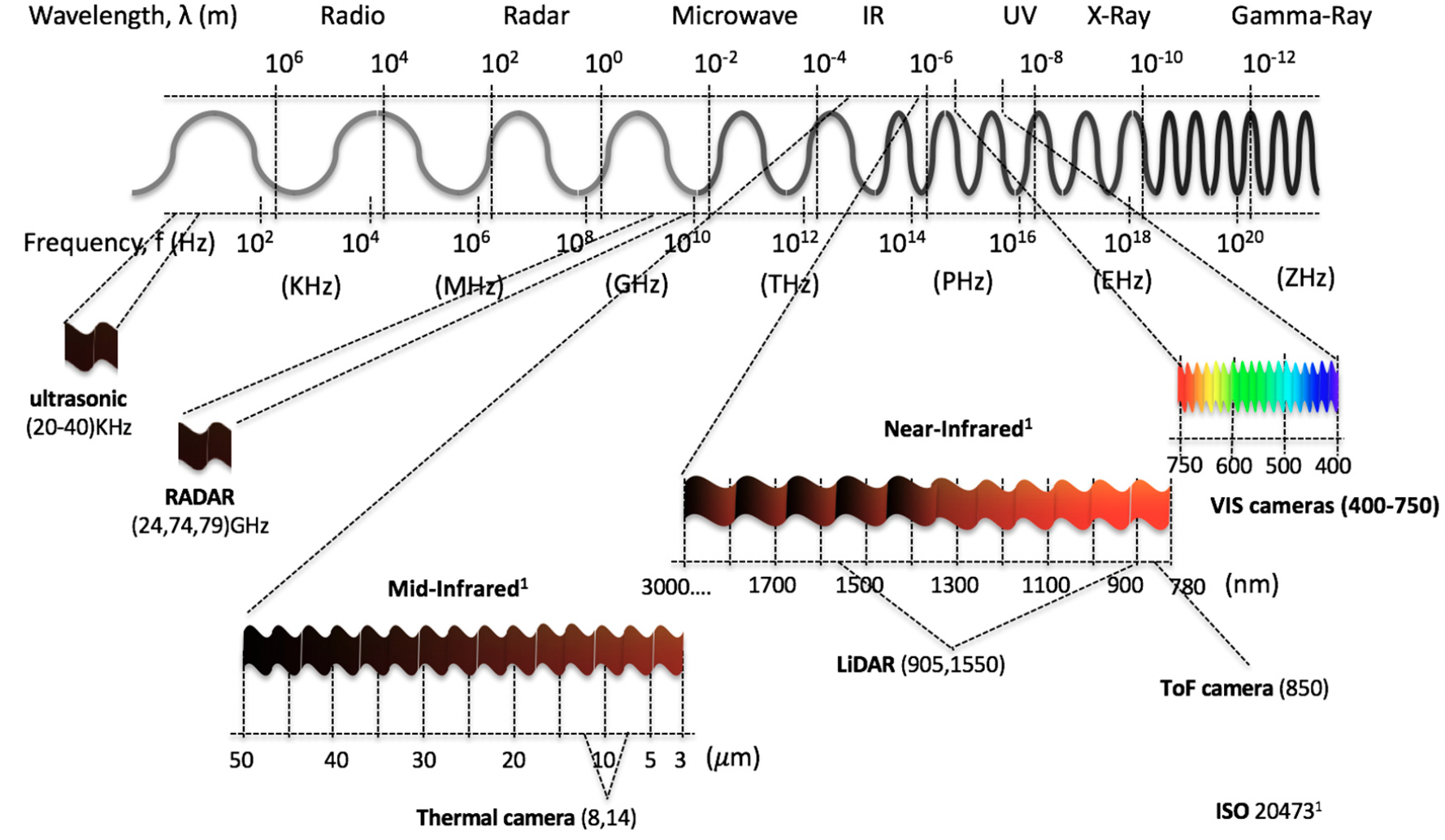}}
    \caption{Spectrum of electromagnetic radiation (Reproduced from \cite{rosique2019systematic})}
    \label{fig:spectrum}
\end{figure}

There are two main methods of conducting ToF measurements in order to calculate object distances. This can be done by measuring the actual time-of-flight of the laser beam based on the time delay between the pulse leaving the emitter and being received by the detector. Another way is to measure the phase shift between the emitted and received signal.

\begin{figure}[htb]
\captionsetup{justification=centering}
    \centering
    \includegraphics[scale=0.5]{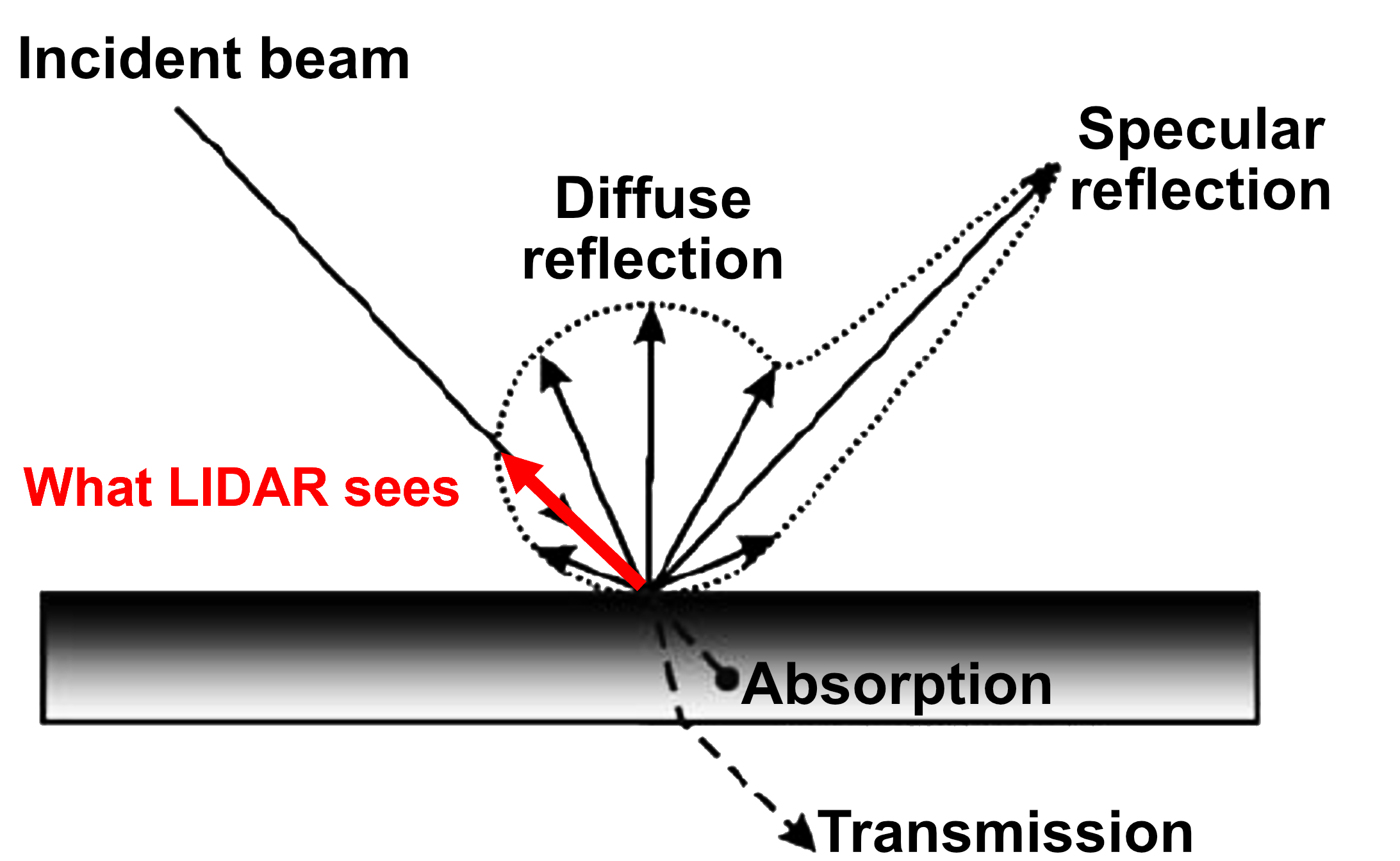}
    \caption{The LiDAR sensor primarily detects the incident reflection.\\At close distances the diffuse and specular reflections may also become visible.}
    \label{fig:LiDAR_reflection}
\end{figure}

Optically, LiDAR works on the principle of diffuse reflection, whereby a ray of light is reflected back at the same angle as the incident beam. This phenomenon is illustrated in Fig.~\ref{fig:LiDAR_reflection}. The LiDAR measures the ToF of a pulse of light from the moment it gets emitted from a laser diode until it is received by a detector after it is reflected by a reflective surface. At very close distances, the LiDAR detectors may also pick up beams from the specular reflection. The range of surrounding objects can then be accurately calculated from these ToF measurements and a point-cloud of the environment is constructed.
\subsubsection{Data outputs}
The data output from LiDARs vary based on manufacturer specifications. Velodyne LiDARs typically return a value known as \textit{calibrated reflectivity} which is described as ``reflectivity values returned based on NIST\cite{NIST}-calibrated reflectivity targets'' from their factory \cite{vlp16_usermanual}. This is nothing but the returned power value of the reflected light as measured against the calibrated reflectivity NIST test targets. Other LiDAR OEMs such as Robosense and Quanergy also return the reflected intensity or reflectivity value along with the X, Y and Z coordinates of each point in space. Sensor manufacturers also implement comprehensive and proprietary post-processing algorithms within the sensors to ensure optimal performance under different lighting conditions.

\subsubsection{Types of LiDAR, advantages and disadvantages}
There are various types of LiDARs available in the market today. Some of the types of LiDARs used for automated driving and Advanced Driver Assistance Systems (ADAS) applications are:
\begin{itemize}
    \item Electromechanical LiDAR
    \item Solid state LiDAR
\end{itemize}

Within each of these categories, there exists both 2D and 3D LiDARs. The following paragraphs will explain different LiDAR technologies, its advantages and disadvantages.

\textbf{Electromechanical LiDAR -} 
Electromechanical LiDARs collect data over a wide area of up to 360 degrees by physically rotating a laser and receiver assembly, or by using a rotating mirror to steer a light beam. Electromechanical LiDARs use powerful, collimated lasers that concentrate the return signal on the detector through highly focused optics \cite{whylidar}.

Electromechanical LiDARs can be designed to return either a 2D or 3D point cloud. In a 2D LiDAR, there may only be a single spinning laser beam or a linear array of laser beams arranged in a single axis. The resulting point cloud may have up to a 360~\degree coverage, but may only provide information about obstacles in a single plane. 2D LiDARs usually provide a low resolution segment-based detection map. This makes them useful for near field applications to detect obstacles within the immediate vicinity of the AV. It is suitable for performing simple detection and ranging tasks on planar surfaces and are not suitable for detailed object detection and all-round perception for any autonomous system \cite{Infographic}.

In a 3D electromechanical LiDAR, the emitter-detector array may be arranged either in a vertical linear fashion, or as a planar array around the rotational axis. High resolution sensors such as the Velodyne VLS-128 have a complex laser emitter-detector pattern in order to accommodate the large number of emitter-detector pairs and to improve scanning resolution. 

\textbf{Solid State LiDAR -}
Solid State LiDARs are designed and built without motorized mechanical scanning. It has no moving mechanical parts and works by providing complete, instantaneous scene illumination through emitted laser flashes and capture incremental insights on objects \cite{whylidar}.

2D solid state LiDARs may have only a linear array of emitter-detector pairs. With advances in MEMS technologies and improved manufacturing techniques, solid state 3D LiDARs are also being developed. These typically rely on a 2-dimensional array of emitter-detectors which emit laser flashes in order to scan a 3D area in front of the sensor and generate the point cloud.

In general, 3D LiDARs are more suitable for detailed analysis of an environment and thus are suited for tasks such as object detection and collision avoidance on automated vehicles \cite{Infographic}. Since they are used for detailed perception tasks, 3D LiDARs typically are designed with the objective of longer detection ranges achieved using higher laser emitter power and detector amplification.

\subsection{Automotive paint}

\subsubsection{Technical properties - pigments/coatings}
Automotive paints specifically refer to the paints used by the automotive industry and applied onto vehicles for the purposes of protecting external body panels and improving the appearance of the vehicle \cite{toda2012automotive}. Automotive paints are usually applied in multiple layers onto the panel of a vehicle body during the production process. It consists of the substrate, which is the panel itself, made of materials such as steel, aluminium, plastics or more recently, composites. A primer is applied to the panel which bonds to it to produce a uniform coating thickness. Next, a base coat, also known as a colour coat is applied onto the panel, giving the surface its colour. Finally, a top coat, also known as a clear coat is applied on top of the base coat to act as a protective coating, mainly to protect the base coat from ultraviolet light damage \cite{toda2012automotive}. 

\subsubsection{Types of automotive paints}
Automotive paints generally fall into categories such as metallic, non-metallic, glossy and matte. The difference between metallic and non-metallic paints lie in the addition of metallic pigments for metallic paints. Metallic pigments consist of very thin platelet-shaped particles made out of aluminum or bronze and they are added on top of the base coat, before applying the top coat. The difference between glossy and matte paints lies in the type of top coat being applied. The top coat can either have a glossy or matte finish. This gives rise to varying abilities of a paint surface to reflect light \cite{automotive_handbk}.
\clearpage

\section{LiDAR and automotive paint testing methodology}
In literature, several studies and experiments \cite{wang2018characterization}, \cite{pomerleau2012noise}, \cite{rasshofer2011influences}, \cite{carrea2016correction}, \cite{reflectivityofpaint} have focused on determining LiDAR performance under varying weather conditions and for simple coloured materials. There have been relatively few experiments focusing specifically on how the paint affects AV perception. This section will focus on the need for testing automotive paint against LiDAR sensors, how does the paint affect AV perception and the LiDAR testing methodology used in our study of \textbf{LiDAR performance against different automotive paint coatings.}

\subsection{What is the need for testing paint against LiDAR?}
Current LiDAR performance as per the Euro NCAP testing standard (for ADAS applications) is assessed on a flat white matte surface with a calibrated reflectivity of 90\%. This is not representative of most vehicles on the road which may be of colours other than white. Vehicles may also possess a highly reflective clear coat or complex matte coatings, resulting in real world performance which cannot be modelled accurately without using real automotive paint samples and representative surfaces. Moreover, testing of automotive LiDAR modules is generally only done either by the OEM in order to parameterize their specifications against specialized test targets or by the end-customers (AV developers) who conduct the testing in order to ensure that the procured sensors satisfy their internal requirements.

In order to address some of these shortcomings, we propose conducting a study to evaluate the impact of automotive paint on LiDAR sensor performance. 

\subsection{How can paint affect AV perception?}
The paint colour and reflectivity of a vehicle body surface usually has an impact on LiDAR based AV perception systems. This is largely due to the ability of the target material to reflect laser light back to the detectors in the LiDAR sensor. In research done to characterise LiDAR sensors for 3D perception, it was found that the reflected intensity of black coloured surfaces is typically lower when compared to a white one \cite{wang2018characterization}.

Additionally, material reflectivity also has an impact on AV perception. Metallic surfaces or material coated with metallic paint presents challenges to LiDAR sensors and thus affects LiDAR based perception. In a study \cite{pomerleau2012noise} to verify the impact of reflectivity, it was established that reflective surfaces such as an aluminium plate poses three challenges to LiDAR sensors. First, when the incident angle is large, most of the energy is not reflected back to the sensor, which can lead to missing measurements. Second, there is also a probability that the laser beam gets reflected to another surface leading to an overestimation of depth. Finally, reflective plates exhibit a larger spectrum of reflected intensity, which seems to create systematic error producing wave patterns. All these sensing challenges could affect AV perception systems, preventing them from carrying out the goal of performing accurate object detection.

\subsection{LiDAR testing methodology}
\subsubsection{Test plan and inputs}
To address the concerns of how paint affects LiDAR visibility performance as mentioned in previous sections, in this white paper we propose a physical LiDAR sensor testing methodology. This will consist of testing LiDAR sensors in controlled environments against common automotive paints and coatings applied on a test target, while varying properties such as range and orientation angle of the test target.

In order to keep variables (such as environmental effects and stray light) to a minimum and ensure controllability and repeatability, the experiments will be conducted under the following environmental conditions:
\clearpage

\begin{itemize}
    \item \textbf{Location} 
    \begin{itemize}
        \item Indoors (In a large indoor laboratory): 40m x 11m x 7m
        \item Outdoors (Mainly for verification testing): 25m x 5.5m
    \end{itemize}
    \item \textbf{Lighting}
    \begin{itemize}
        \item Office fluorescent lighting (Indoors)
        \item Sunlight (Outdoors)
    \end{itemize}
\end{itemize}

There are two primary test inputs to be considered when conducting the experiment. They are:
\begin{itemize}
    \item \textbf{Sensor under test} - It is important to test multiple LiDAR sensors with varying designs and properties in order to gain a thorough understanding of the challenges faced by AV sensing and perception systems against other vehicles and obstacles on the road. The LiDARs that would be part of the experimental setup are as follows:
    \begin{itemize}
        \item \textbf{Electromechanical LiDAR
       } \begin{enumerate}
            \item Velodyne VLS-128 (128 channel, long range)
       
       \item Velodyne VLP-16 (16 channel, short to medium range)
         \end{enumerate}
        \item \textbf{Solid-state LiDAR}
        \begin{enumerate}
       \item Leddartech Pixell (8 vertical / 96 horizontal channels, short to medium range)\end{enumerate}
    \end{itemize}
    \item \textbf{Panel under test} - Automotive paints of both non-metallic and metallic finishes will be tested. These sample paint panels are obtained through a partnership with paint and coating supplier NIPSEA Technologies, Nippon Paint's Southeast Asia R\&D division. The paint panels supplied were recommended by NIPSEA and are based on a selection of paints representative of the vehicles in Singapore. The paints were chosen as per table \ref{table:Paint_colours}. The complete list of test panels along with their material properties and code names are listed in table \ref{table:NIPSEA_paints}.

The sample paint panel under test also needs to be tested with different parameters and
physical states as highlighted below:

\begin{enumerate}
\item \textbf{Varying elevation angle} - From 0\degree~to 70\degree~at 5\degree~intervals
    \begin{itemize}     
    \item 0\degree~being normal to the LiDAR beams  
    \end{itemize}
\item \textbf{Varying azimuth angle} - From 0\degree~to 70\degree~at 5\degree~intervals (For verification testing) 
    \begin{itemize}     
    \item This is done by tilting the test panels by a 90\degree~angle  \end{itemize}   
\item
\textbf{Varying target distance} - From 2.5m to 30.0m at selected distances
\begin{itemize}     
    \item Velodyne VLS-128: 2.5m, 5.0m, 10.0m, 30.0m
    \item Velodyne VLP-16 \& Leddartech Pixell: 2.5m
    \end{itemize}
    \end{enumerate}

Due to the Velodyne VLP-16 and Leddartech Pixell being lower resolution and lower powered LiDARs, they have limited operating range as compared to the VLS-128. Hence, to measure a sufficiently large number of points within the limited test panel size, distance readings for both LiDARS were only taken up to 2.5m.

\end{itemize}
\clearpage

\begin{table}[htb]
\centering
\begin{tabular}{|l|l|l|l|}
\hline
No. & Colour & Finish          & Colour description                    \\ \hline
1   & Black  & Gloss / matt    & -                                     \\ \hline
2   & White  & Gloss           & -                                     \\ \hline
3   & Blue   & Gloss           & Similar to Comfort DelGo taxi colours \\ \hline
4   & Red    & Gloss, metallic & Similar to TransCab taxi colours      \\ \hline
5   & Green  & Gloss           & Similar to SGBus colours              \\ \hline
6   & Silver & Gloss, metallic & Similar to Toyota's standard colour   \\ \hline
\end{tabular}%
\caption{A summary of the paint colours chosen}
\label{table:Paint_colours}
\end{table}

\begin{table}[htb]
\centering
\begin{tabular}{|l|c|c|l|l|}
\hline
No. & Colour                                                                       & \multicolumn{1}{l|}{Surface Finish} & Panel Code  & Remarks              \\ \hline
1   & \multirow{5}{*}{Black}                                                       & Gloss                               & SB-Gloss*    &                      \\ \cline{1-1} \cline{3-5} 
2   &                                                                              & Gloss                               & FB1-Gloss**   & LiDAR functionalised \\ \cline{1-1} \cline{3-5} 
3   &                                                                              & Matte                               & SB-Matt*     &                      \\ \cline{1-1} \cline{3-5} 
4   &                                                                              & Matte                               & FB1-Matt**    & LiDAR functionalised \\ \cline{1-1} \cline{3-5} 
5   &                                                                              & Gloss                               & FB4-Gloss**   & LiDAR functionalised \\ \hline
6   & White                                                                        & Gloss                               & SW-Gloss    &                      \\ \hline
7   & \multirow{2}{*}{Blue}                                                        & Gloss                               & CDSBL-Gloss &                      \\ \cline{1-1} \cline{3-5} 
8   &                                                                              & Gloss                               & CDFBL-Gloss & LiDAR functionalised \\ \hline
9   & \multirow{2}{*}{\begin{tabular}[c]{@{}c@{}}Red\\ (Metallic)\end{tabular}}    & Gloss                               & TCSRM-Gloss &                      \\ \cline{1-1} \cline{3-5} 
10  &                                                                              & Gloss                               & TCFRM-Gloss & LiDAR functionalised \\ \hline
11  & Green                                                                        & Gloss                               & SMRTG-Gloss &                      \\ \hline
12  & \multirow{2}{*}{\begin{tabular}[c]{@{}c@{}}Silver\\ (Metallic)\end{tabular}} & Gloss                               & TSSM-Gloss  &                      \\ \cline{1-1} \cline{3-5} 
13  &                                                                              & Gloss                               & TFSM-Gloss  & LiDAR functionalised \\ \hline
\end{tabular}
\medskip
\medskip
\\
*SB / Standard black -- Black paint with standard black pigment\\
**FB / Functionalised black -- Variants of black paint with special LiDAR functionalised pigment
\caption{A summary of the tested paint panels from NIPSEA}
\label{table:NIPSEA_paints}
\end{table}

\subsubsection{Overview of proposed LiDAR testing setup}
LiDAR sensors perform best indoors without errant reflections or interference from sunlight.
Therefore, the proposed test setup is primarily located indoors under controlled fluorescent lighting. Fluorescent
lighting does not affect LiDAR sensors typically since they do not emit in the IR spectrum. Verification tests were done to confirm this and the results can be found in the subsequent sections.

\begin{figure}[htb]
    \centering
    \centerline{
        \includegraphics[scale=0.6]{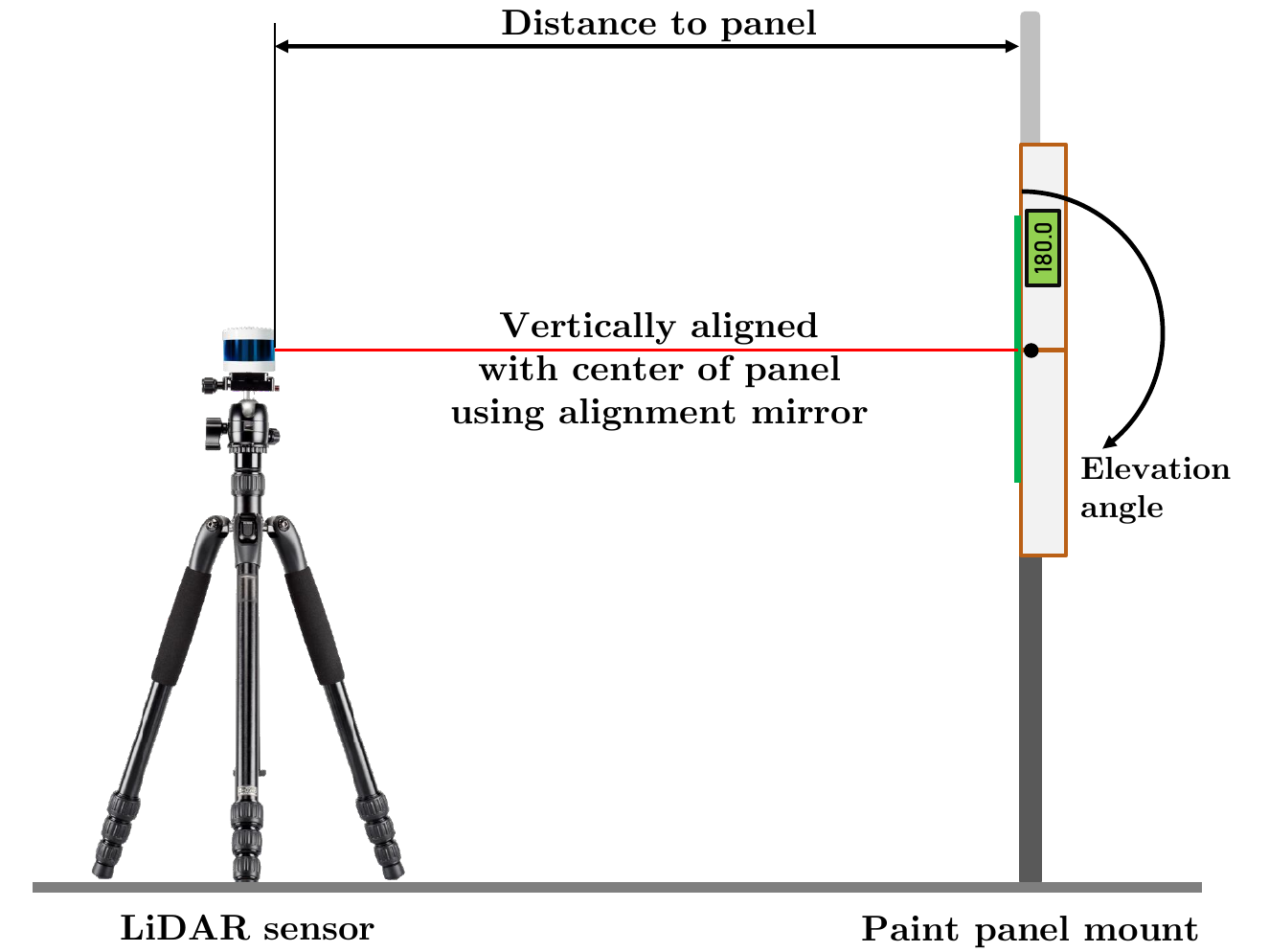}
    }
    \caption{Proposed LiDAR testing setup}
    \label{fig:test_setup}
\end{figure}

The test setup consists of the following as illustrated in Fig.~\ref{fig:test_setup}:
\begin{itemize}
    \item \textbf{Sensor under test} - The LiDAR sensor is mounted  on a height adjustable tripod. The height of the LiDAR sensor matches that of the panel mount and its alignment mirror as described subsequently.
    \item \textbf{Panel under test} - Sample paint panels were obtained through a partnership with paint and coating supplier NIPSEA Technologies.  These sample panels are flat square nickel coated mild steel plates and are spray-painted manually by the supplier.
    \begin{itemize}
        \item Dimensions for a single panel are 50cm x 50cm due to the physical constraints of the manual spray booth
        \item To test larger targets, 4 pieces can be combined to obtain a 1m x 1m test target
        \item Black colour LiDAR functionalised panels consist of 2 variants (FB1 and FB4). They differ in the type of base coat being applied, by adjusting the paint's functional black colour coating system and calibrating it to achieve different Near Infrared Radiation (NIR) Reflectance values, this in turn affects LiDAR visibility of the 2 variants. 
        \item Standard White (SW-Gloss) and SMRT Green (SMRTG-Gloss) panels have no LiDAR functionalised version because the paint colour itself does not contain any amount of black colour, so the LiDAR functionalised pigments which are black cannot be added.
    \end{itemize}
     \item \textbf{Panel mount} - The panel is mounted with a pivoting setup with adjustable elevation angles. The pivoting setup is developed from standard white board hardware with specific modifications made to improve test reliability and repeatability:
    \begin{itemize}
        \item The elevation angle can be measured using a digital angle gauge which is physically and securely attached to the pivot points of the test rig.
        \item Reference reflectivity is measured using matte black paper and the white board's own gloss white painted surface (the whiteboard has a white gloss paint surface on a steel substrate).
        \item The panel under test is held in place on the white board surface using carefully positioned and aligned high-strength neodymium magnets. These magnets help to align the panels and also hold them flat against the surface ensuring that the panels are perfectly flat.
        \item The white board surface also includes a mirror located at the centre of the pivot point. This mirror helps to align the sensor with the white board surface. When the sensor is perfectly aligned, the point-cloud will display a small region with maximum intensity. This alignment method is accurate to 0.1\degree~in terms of elevation/azimuth angles and 1cm in terms of translation.
        \item In order to vary the azimuth angle of the paint panels for verification testing, the entire panel mount together with the mounted test panels will be tilted by 90\degree, so that the panel mount lies on its side. This enables the azimuth angle of the panels to be varied without having to modify the panel mount.
    \end{itemize}
    \item \textbf{LiDAR data capture computer} - A computer is connected to the LiDAR to capture in real-time the output point-cloud in order to analyze sensor performance against various test targets. The data capture computer runs ROS with custom-developed packages in order to automate measurement.
\end{itemize}

\begin{figure}[htb]
    \centering
    \centerline{
        \includegraphics[scale=0.5]{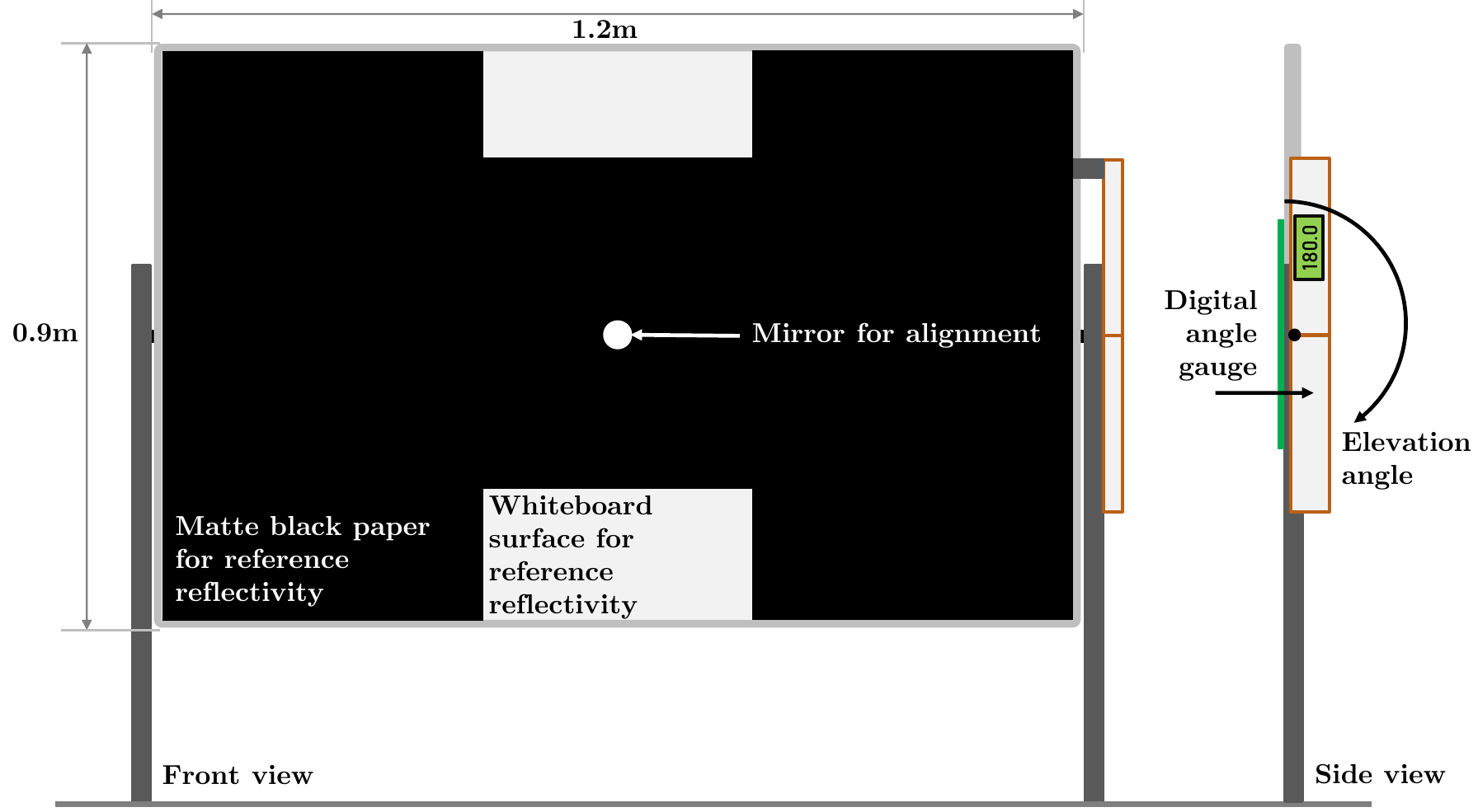}
    }
    \caption{Customised whiteboard for mounting the test paint panels, front and side view}
    \label{fig:Whiteboard_1}
\end{figure}

\begin{figure}[htb]
    \centering
    \centerline{
        \includegraphics[scale=0.5]{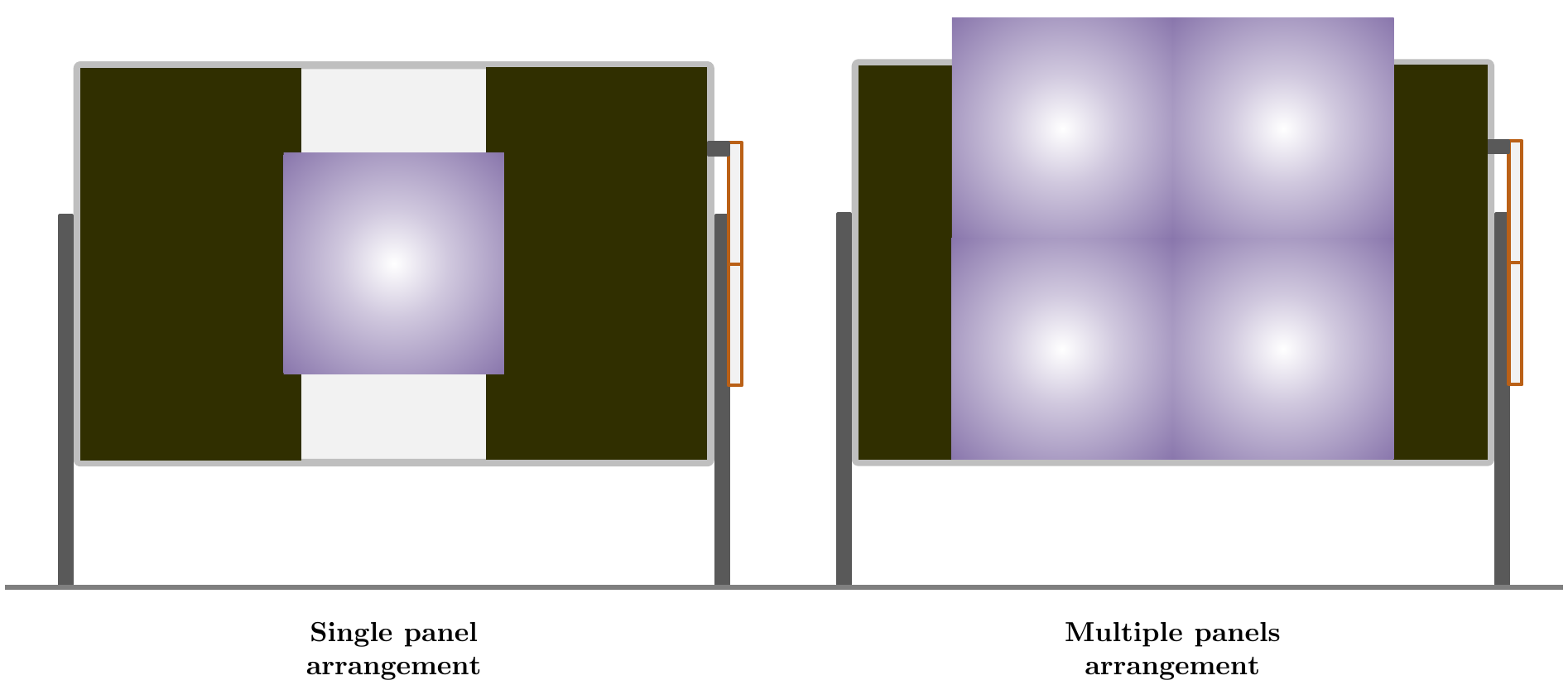}
    }
    \caption{Panel mount with single and multiple panels}
    \label{fig:Whiteboard_3}
\end{figure}

\begin{figure}[htb]
\captionsetup{justification=centering}
    \centering
    \centerline{
        \includegraphics[scale=0.5]{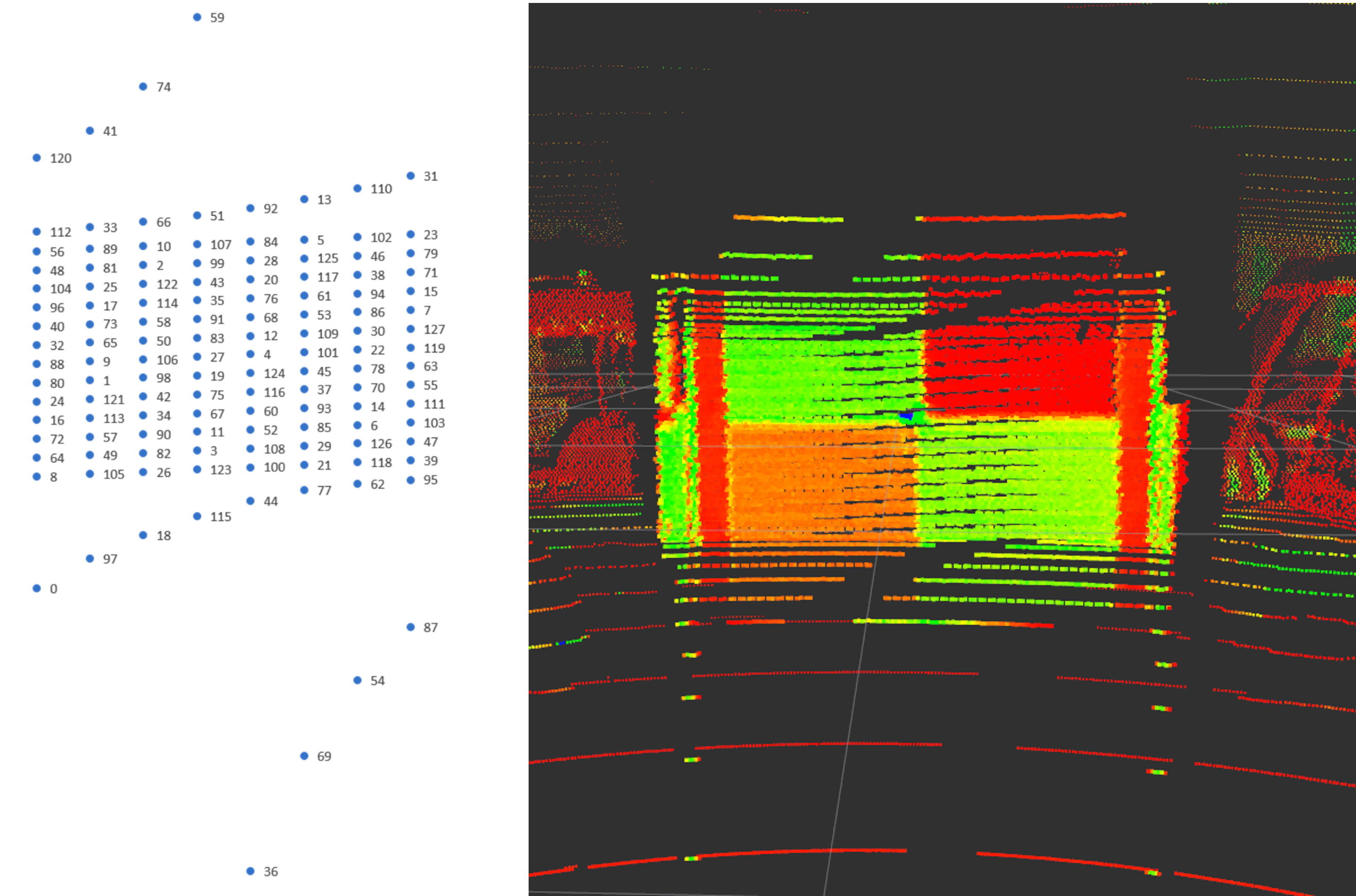}
    }
    \caption{Beam pattern of Velodyne VLS-128 and corresponding point cloud frame.\\This beam pattern is only seen at extremely low rotational speeds. Hence, it does not affect the test readings.}
    \label{fig:VLS_128_beam}
\end{figure}

\clearpage
\section{Results and analysis}
The LiDAR testing was done with a quantitative analysis of panel reflectivity using the three LiDARs as mentioned earlier in the \textit{Sensor under test} subsection. They are namely the Velodyne VLS-128, Velodyne VLP-16 and Leddartech Pixell with each LiDAR configuration detailed below:
\paragraph{LiDAR Sensors}
\begin{itemize}
\item \textbf{Velodyne VLS-128} - in `Strongest return' mode:
    \begin{itemize}
        \item Horizontal FoV: 360\degree
        \item Vertical FoV: -25\degree ~to +15\degree, non-linear beam pattern as seen in Fig.~\ref{fig:VLS_128_beam}
        \item Rotation speed: 540 rpm
        \item Mount position: Horizontally mounted
    \end{itemize}
    \item \textbf{Velodyne VLP-16} - in `Strongest return' mode:
    \begin{itemize}
        \item Horizontal FoV: 360\degree
        \item Vertical FoV: -15\degree ~to +15\degree in 2\degree ~increments
        \item Rotation speed: 600 rpm
        \item Mount position: Horizontally mounted
    \end{itemize}
    \item \textbf{Leddartech Pixell}:
    \begin{itemize}
        \item Horizontal FoV: 177.5\degree
        \item Vertical FoV: Approx. -8\degree ~to +8\degree ~in 2\degree ~increments
        \item Rotation speed: Not Applicable
        \item Mount position: Horizontally mounted
    \end{itemize}
\end{itemize}

\paragraph{Test targets -} 
Automotive grade paint panels were used as test targets. These are metal panels painted with automotive paint and affixed to the panel mount while being held in place by its magnets. The target specifications are listed and described in Table \ref{table:NIPSEA_paints}. 

\begin{figure}[htb]
    \centering
    \centerline{
        \includegraphics[scale=0.5]{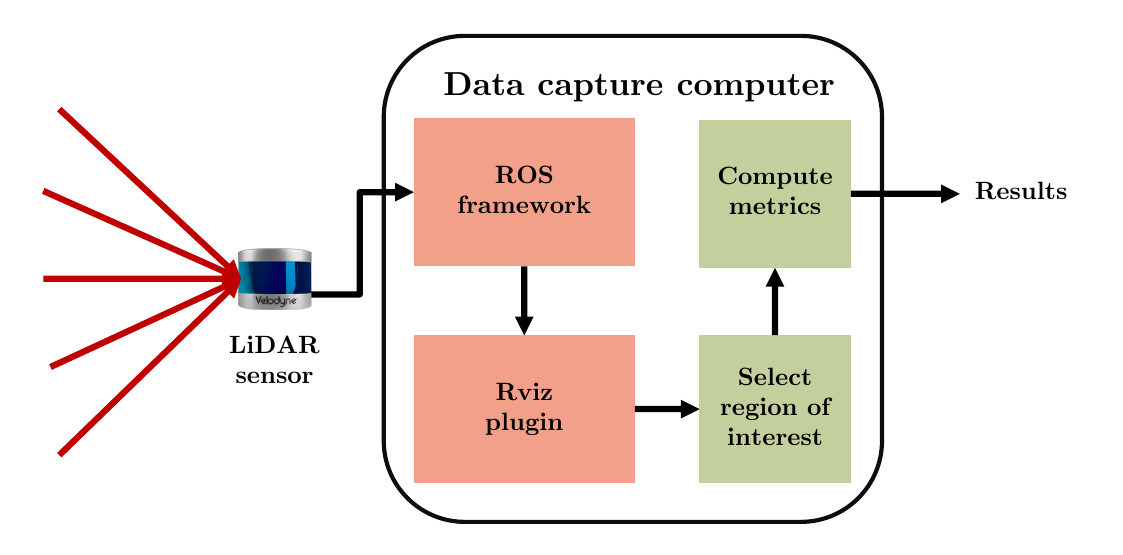}
    }
    \caption{Overview of the data capture software framework}
    \label{fig:Data_capture}
\end{figure}

\paragraph{LiDAR data capture computer -} 
The test results were recorded using the LiDAR sensor data capture computer and its ROS-based framework (Fig~\ref{fig:Data_capture}). The average intensity of the test panel region was then extracted from the point cloud. This ranges from 0 -- 255 for the Velodyne LiDARs and 0 -- 262143 for the Leddartech Pixell.

\paragraph{Parameters -} 
The test parameters were varied for each LiDAR and they are as follows:
\begin{itemize}
\item \textbf{Velodyne VLS-128}

        \begin{itemize}
                \item Selected distances to target: 2.5m, 5.0m, 10.0m, 30.0m
                \item Elevation angles of target at each distance: From 0\degree ~to 70\degree ~at 5\degree ~intervals
                \item Selected test targets: All paint panels as listed in Table \ref{table:NIPSEA_paints}
                \item Conditions: Indoor with lights on
            \end{itemize}

\item \textbf{Velodyne VLP-16}
\begin{itemize}
                \item Selected distances to target: 2.5m
                \item Elevation angles of target at each distance: From 0\degree ~to 70\degree ~at 5\degree ~intervals
                \item Selected test targets: All paint panels as listed in Table \ref{table:NIPSEA_paints}
                \item Conditions: Indoor with lights on
            \end{itemize}
\item \textbf{Leddartech Pixell}:
    \begin{itemize}
        \item Selected distances to target: 2.5m
                \item Elevation angles of target at each distance: From 0\degree ~to 70\degree ~at 5\degree ~intervals
                \item Selected test targets: All paint panels as listed in Table \ref{table:NIPSEA_paints}
                \item Conditions: Indoor with lights on
    \end{itemize}
\end{itemize}

\subsection{Results}

The relevant panels for each test were mounted on the panel mount separately in order to obtain reflected intensity. Three readings were taken for each type of panel and the average reflected intensity was computed. This was done at the selected distances and for multiple angle orientations as mentioned above. The results from the tests were consolidated into average reflected intensity vs angle of incidence graphs. These are further elaborated in the sections below.

\begin{figure}[htb]
    \centering
    \centerline{
        \includegraphics[scale=0.075]{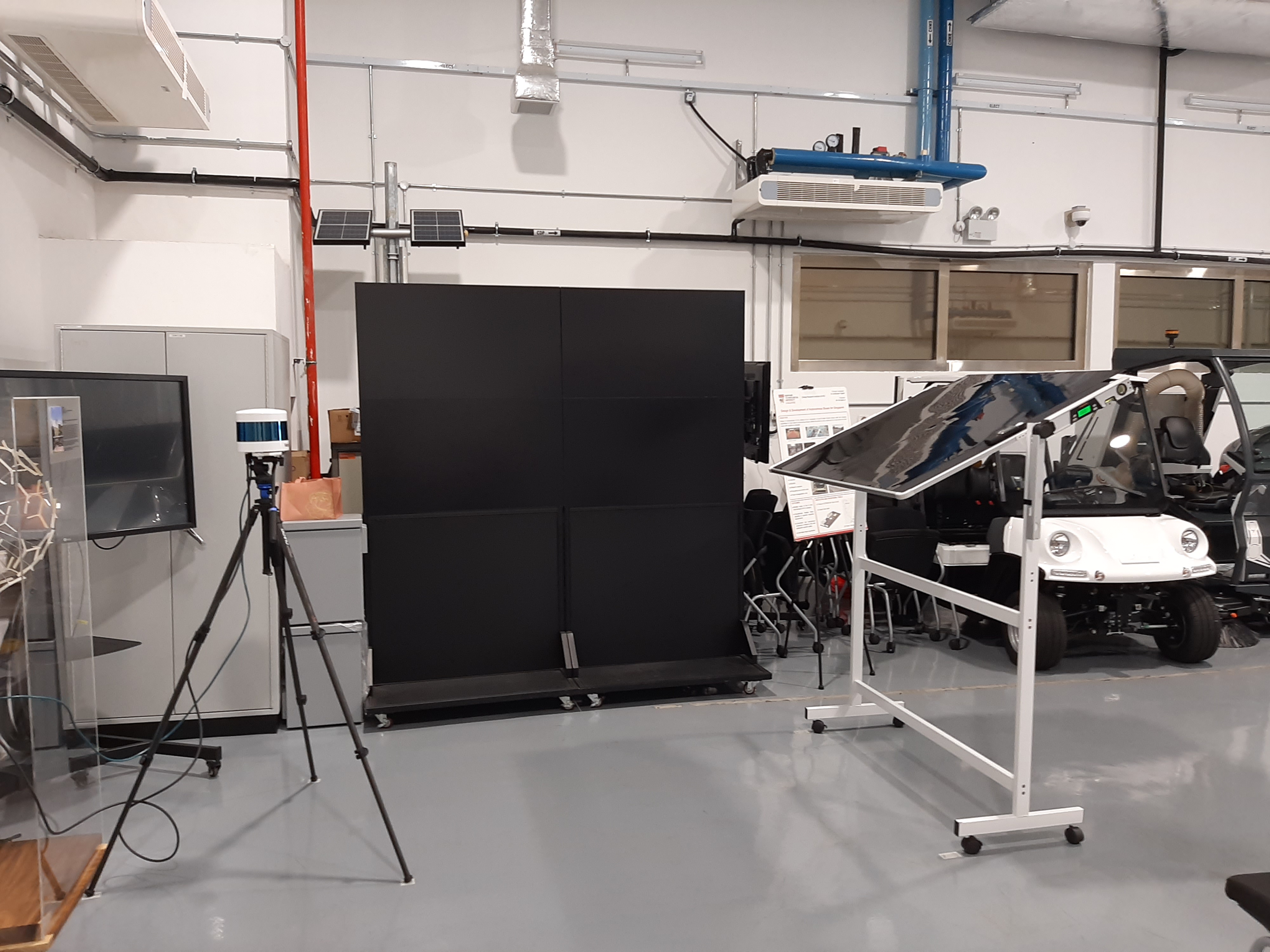}
    }
    \caption{Actual setup of equipment (LiDAR and Panel Mount) in the lab}
    \label{fig:lab_setup}
\end{figure}

\begin{figure}[h!]
    \centering
	\captionsetup{justification=centering}
	\subfloat[]{\label{fig:SB_gloss0} \includegraphics[width=0.26\columnwidth]{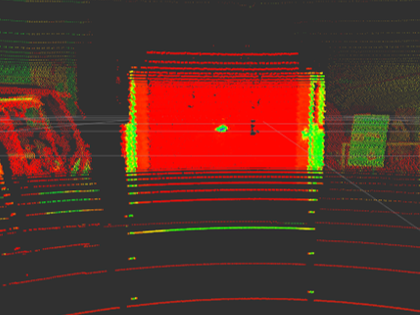}}
	\hspace{1pt}
	\subfloat[]{\label{fig:SB_gloss50} \includegraphics[width=0.26\columnwidth]{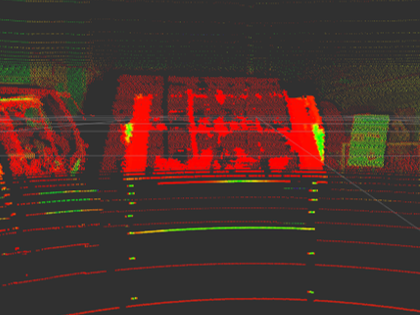}}
	\hspace{1pt}
	\subfloat[]{\label{fig:SB_gloss70} \includegraphics[width=0.26\columnwidth]{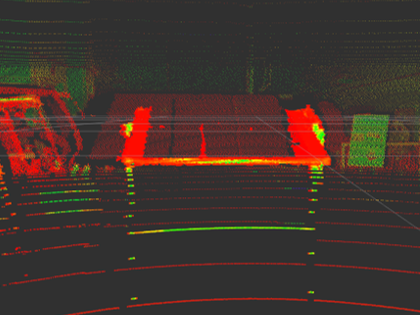}}
    \caption{Sample SB\_Gloss point cloud at varying panel angles of (a) 0\degree, (b) 50\degree~and (c) 70\degree}
    \label{fig:SB_gloss}
\end{figure}

\subsubsection{Tests using Velodyne VLS-128}
\begin{figure}[h!]
    \centering
	\captionsetup{justification=centering}
	\subfloat[]{\label{fig:non-metallic_128} \includegraphics[width=0.48\columnwidth]{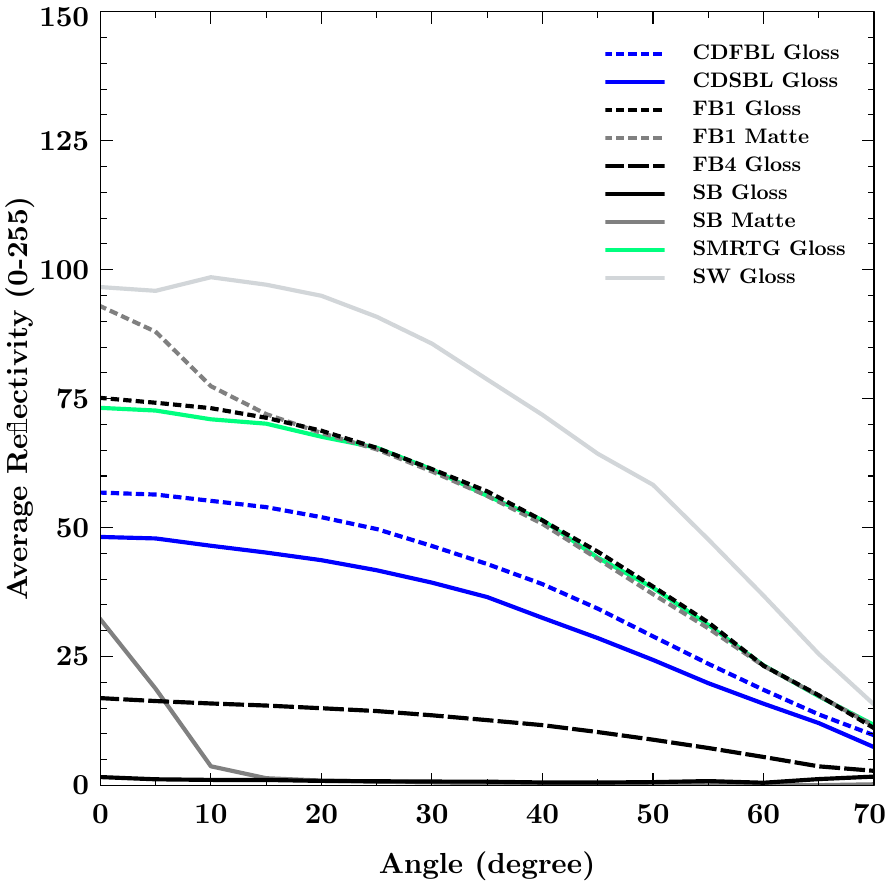}}
	\hspace{1pt}
	\subfloat[]{\label{fig:metallic_128} \includegraphics[width=0.48\columnwidth]{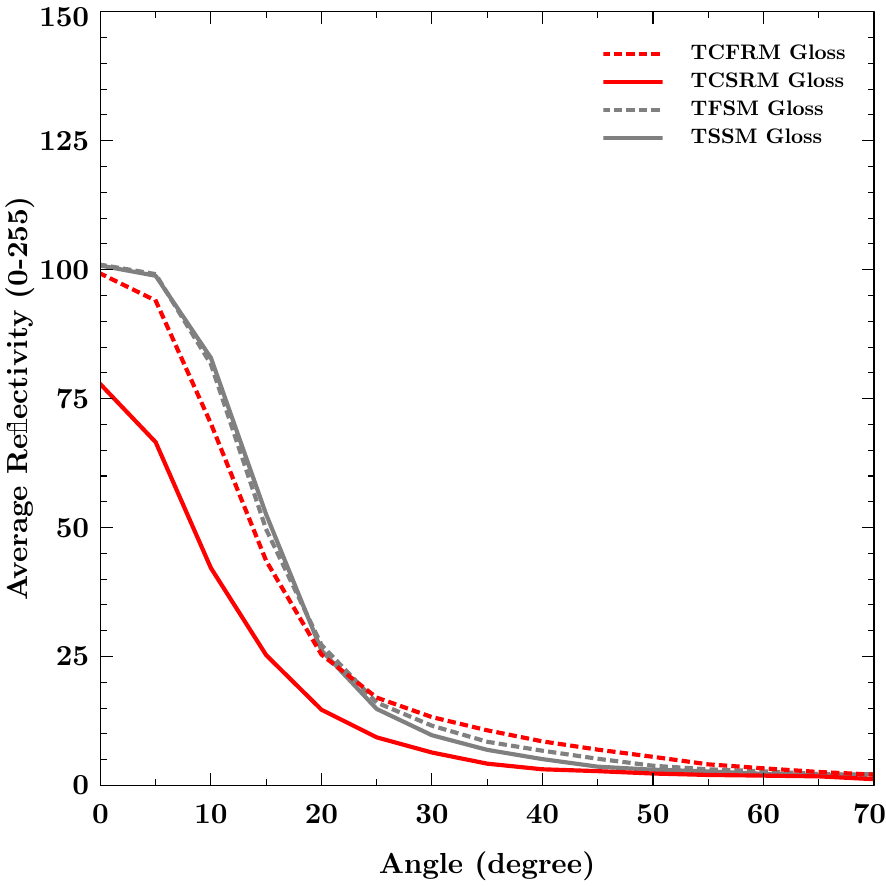}}
	\caption{VLS-128 Average reflected intensity vs. Angle of incidence at 2.5m \textbf{(a)} Non-metallic   \textbf{(b)} Metallic paints}
	\label{fig:VLS-128_2.5m}
\end{figure}

\begin{figure}[h!]
    \centering
	\captionsetup{justification=centering}
	\subfloat[]{\label{fig:non-metallic_128_5m} \includegraphics[width=0.48\columnwidth]{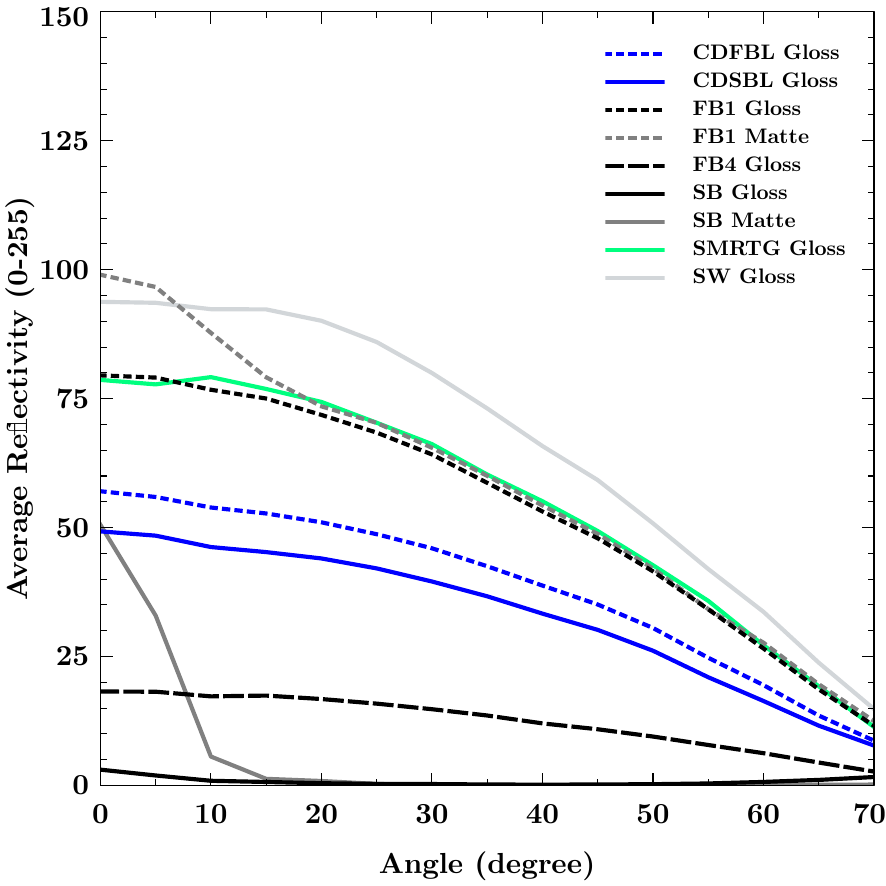}}
	\hspace{1pt}
	\subfloat[]{\label{fig:metallic_128_5m} \includegraphics[width=0.48\columnwidth]{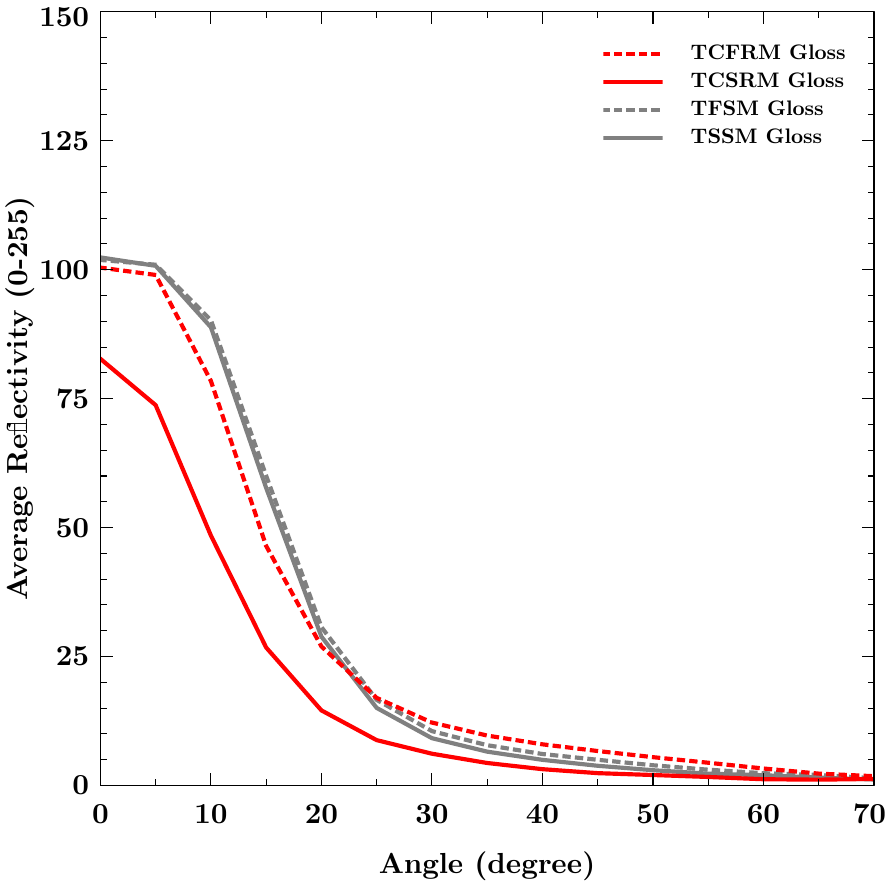}}
	\caption{VLS-128 Average reflected intensity vs. Angle of incidence at 5m \textbf{(a)} Non-metallic   \textbf{(b)} Metallic paints}
	\label{fig:VLS-128_5m}
\end{figure}

\begin{figure}[h!]
    \centering
	\captionsetup{justification=centering}
	\subfloat[]{\label{fig:non-metallic_128_10m} \includegraphics[width=0.48\columnwidth]{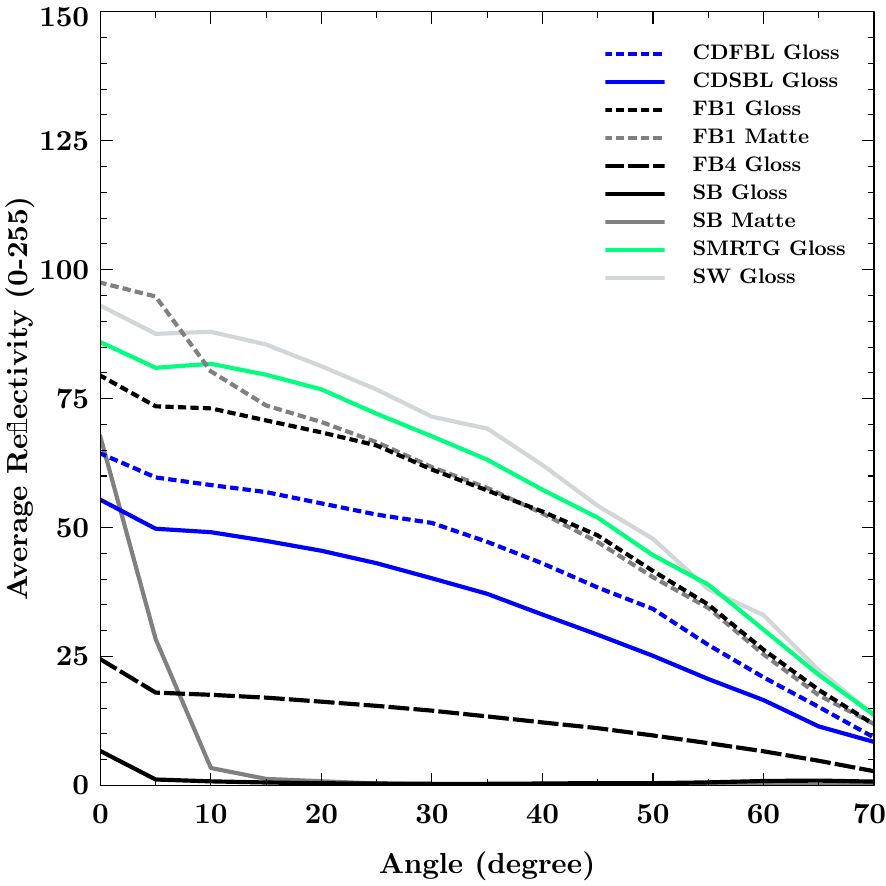}}
	\hspace{1pt}
	\subfloat[]{\label{fig:metallic_128_10m} \includegraphics[width=0.48\columnwidth]{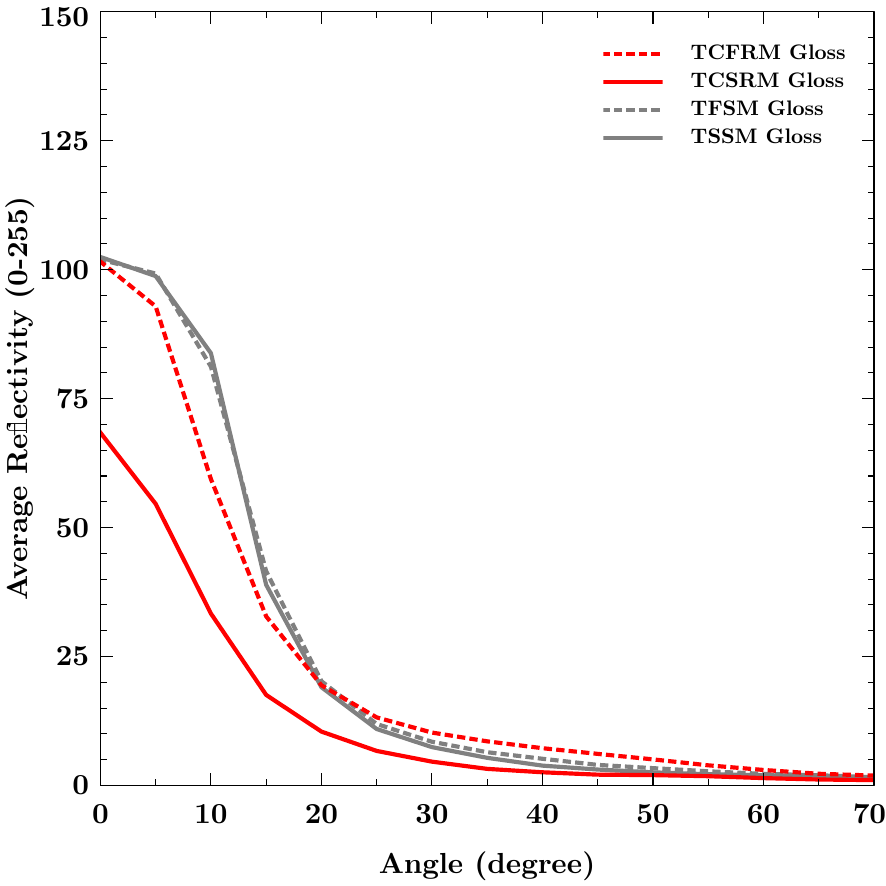}}
	\caption{VLS-128 Average reflected intensity vs. Angle of incidence at 10m \textbf{(a)} Non-metallic   \textbf{(b)} Metallic paints}
	\label{fig:VLS-128_10m}
\end{figure}

\begin{figure}[h!]
    \centering
	\captionsetup{justification=centering}
	\subfloat[]{\label{fig:non-metallic_128_30m} \includegraphics[width=0.48\columnwidth]{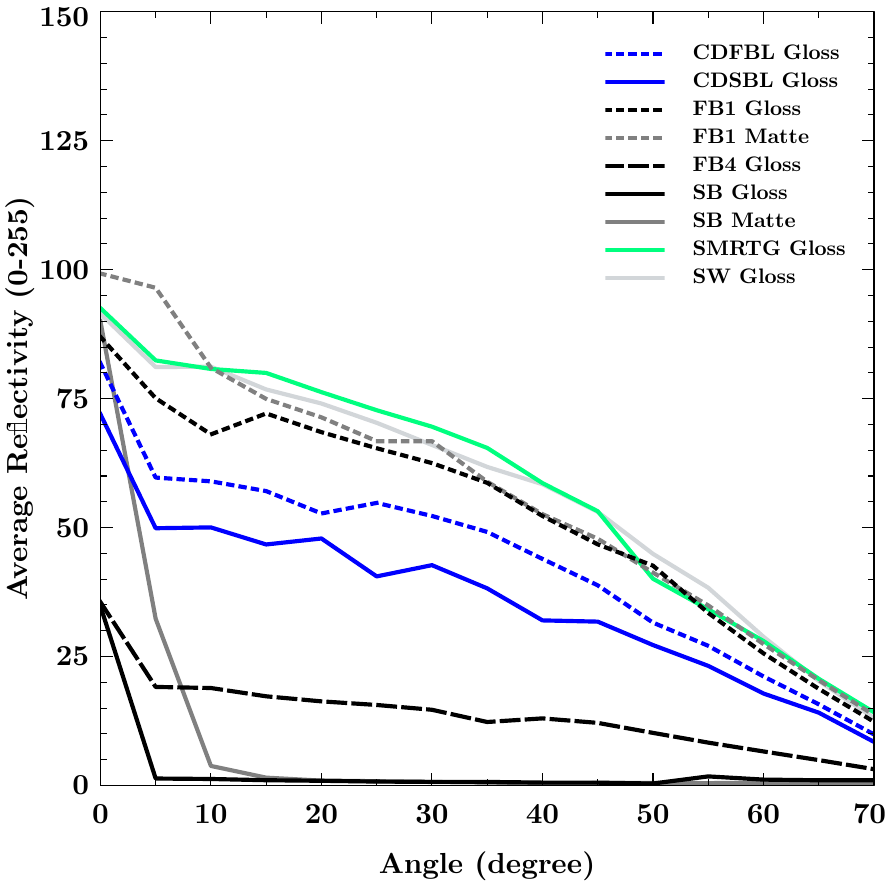}}
	\hspace{1pt}
	\subfloat[]{\label{fig:metallic_128_10m} \includegraphics[width=0.48\columnwidth]{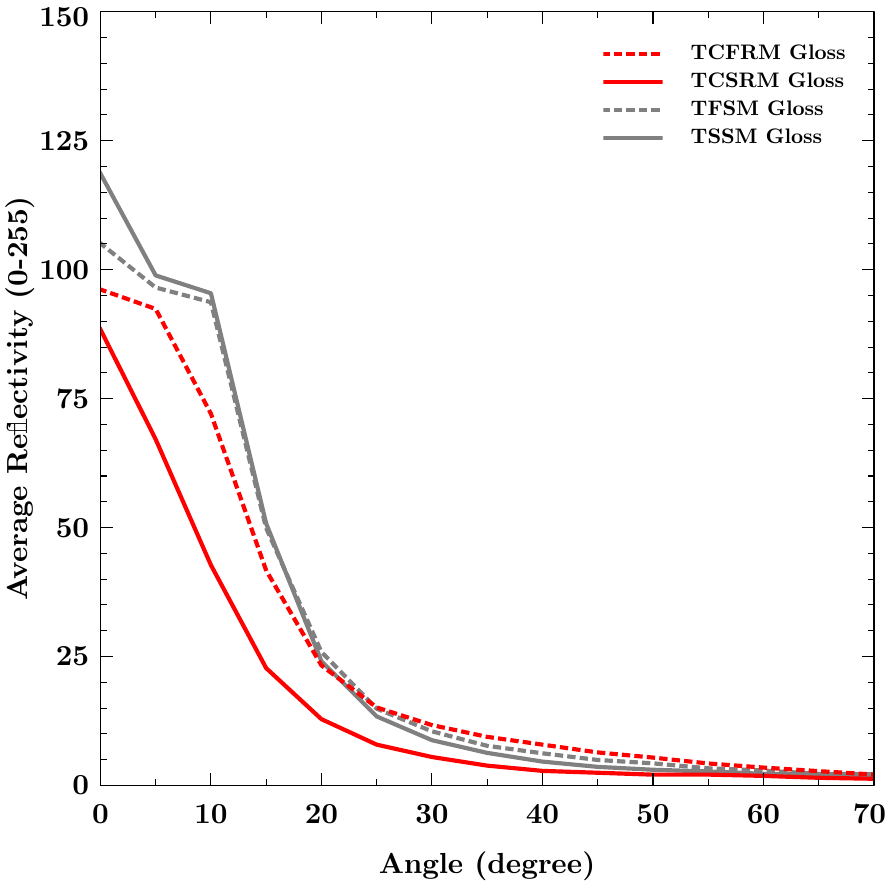}}
	\caption{VLS-128 Average reflected intensity vs. Angle of incidence at 30m \textbf{(a)} Non-metallic   \textbf{(b)} Metallic paints}
	\label{fig:VLS-128_30m}
\end{figure}

From the experimental results across all selected distances, SB\_Gloss shows the worst LiDAR reflection performance and SW\_Gloss is able to maintain very high LiDAR reflectivity as the angle increases. This is illustrated in Fig. \ref{fig:VLS-128_2.5m}, \ref{fig:VLS-128_5m}, \ref{fig:VLS-128_10m}, \ref{fig:VLS-128_30m} which are the graphs of average reflected intensity vs. angle of incidence at selected distances. SB\_Gloss is a non-metallic black shade typically present on road vehicles and these results show that LiDAR sensors have significant difficulty seeing black colour. The results also show that LiDAR sensors are able to see white colour better than most other colours as evidenced by the results for SW\_Gloss which is a non-metallic white shade. There is also a noticeable slight increase in intensity for the SW\_Gloss at 5\degree ~when the panels are placed at distance of 2.5m. According to the paint supplier, this could be due to internal coating reflections within the paint panel. However, black colour LiDAR functionalised paint types, namely FB1 and FB4 gloss and matt variants have significantly improved reflected intensities when compared to the SB\_Gloss panel.

Moreover, we observe that there are two groups of paints with distinct intensity vs angle trends. This can also be seen at all selected distances. One group consist of paint types (metallic paints) where the intensity drops sharply as the angle increases. Another group consists of paint types (non-metallic paints) where intensity holds up better and decreases gradually as the angle increases. Even though the trend seen for the matte paints may be due to different reasons as compared to the metallic paints, a large specular reflection spot was observed for the SB\_Matt panel at 0\degree ~which completely disappeared at 15\degree.
\clearpage

\clearpage

\clearpage

\begin{figure}[h!]
    \centering
    \centerline{
        \includegraphics[scale=0.52]{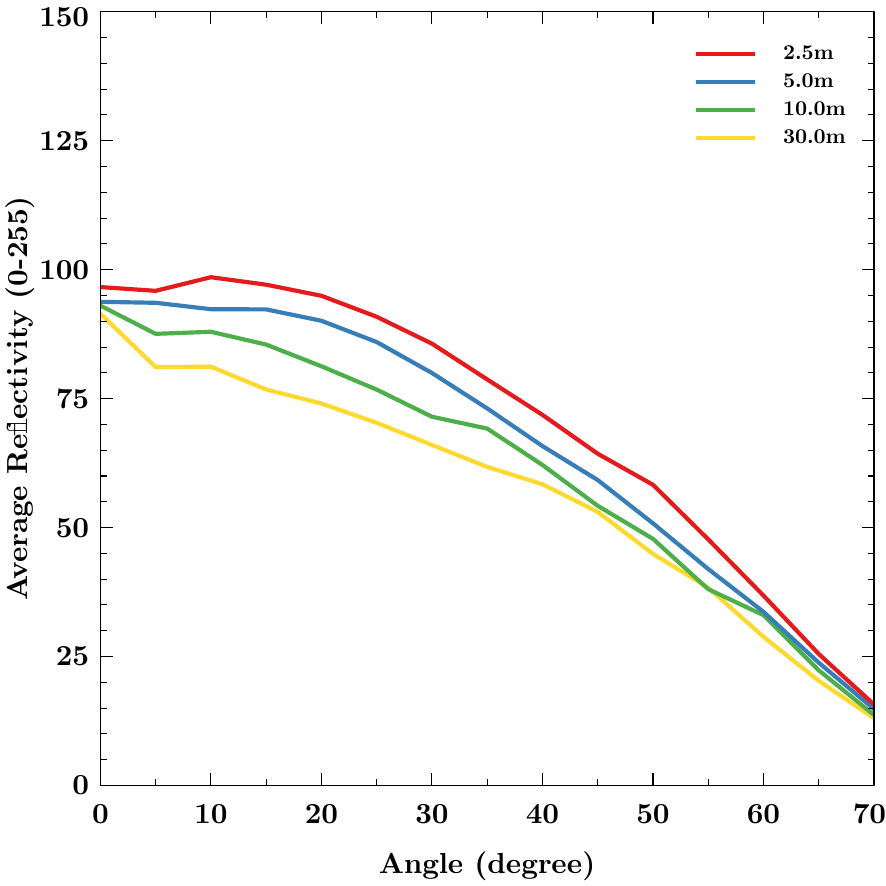}
    }
    \caption{VLS-128 Average reflected intensity vs. Angle of incidence for SW\_Gloss panel at selected distances}
    \label{fig:SWGloss_distance}
\end{figure}

For visualization of the effects of distance on reflected intensity, a graph comparing the SW\_Gloss panel data at selected distances of 2.5m, 5m, 10m and 30m was created in Fig. \ref{fig:SWGloss_distance}. The figure shows that intensity decreases slightly as panel distance from the LiDAR increases. This slight drop in intensity is likely due to the loss of energy in the laser beams emitted from the LiDAR as it travels through the atmosphere.

\begin{figure}[h!]
    \centering
    \centerline{
        \includegraphics[scale=0.52]{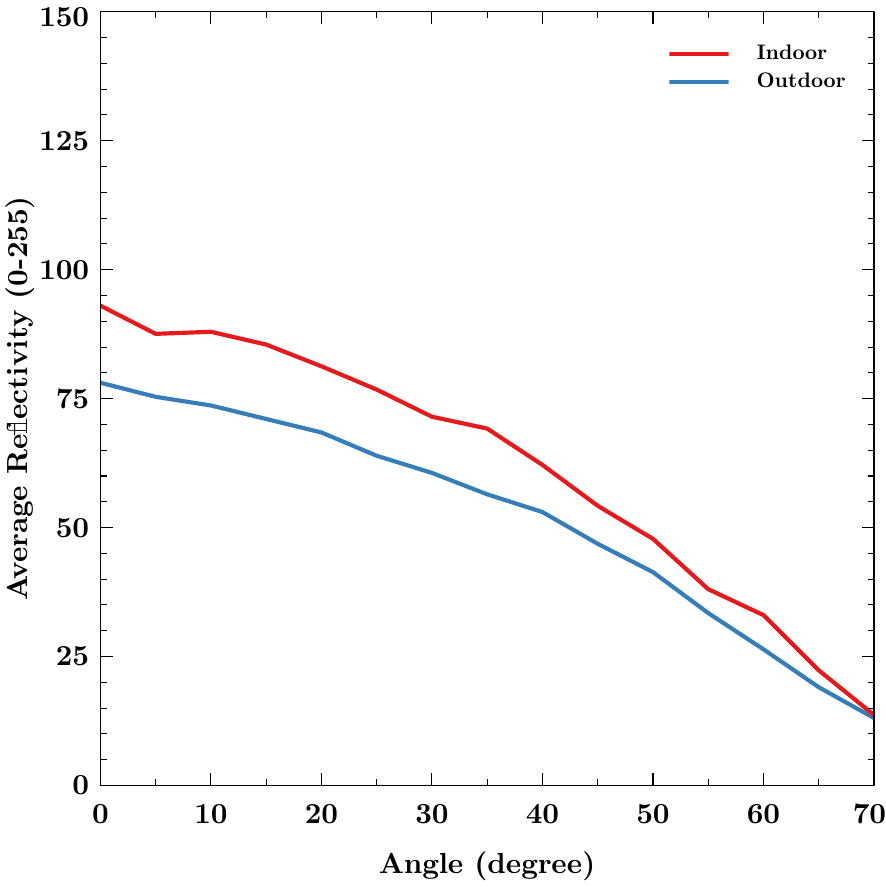}
    }
    \caption{VLS-128 Average reflected intensity vs. Angle of incidence for SW\_Gloss panel at 10m different locations}
    \label{fig:SWGloss_outdoor}
\end{figure}

\textbf{Outdoor Verification Tests}\\
Outdoor verification testing was carried out with the VLS-128 at a large outdoor space of around 25m x 5.5m. Three different panel types (SW\_Gloss, SB\_Gloss and SB\_Matt) were placed at a distance of 10m from the LiDAR and their average return intensities were recorded. The data collected outdoor was then compared with those collected indoor. A comparison of indoor and outdoor data for the SW\_Gloss panel is shown in Fig. \ref{fig:SWGloss_outdoor} as a sample of the entire data. From the figure, we can see that the indoor and outdoor trend is similar, with both showing a gradual decrease in intensity as angle increases for the SW\_Gloss panel. Hence, the results show that performing the tests indoors is still a valid representation of how a LiDAR sensor would perform when it is mounted on a vehicle outdoors.

\clearpage

\subsubsection{Tests using Velodyne VLP-16}

\begin{figure}[h!]
    \centering
	\captionsetup{justification=centering}
	\subfloat[]{\label{fig:non-metallic_16} \includegraphics[width=0.48\columnwidth]{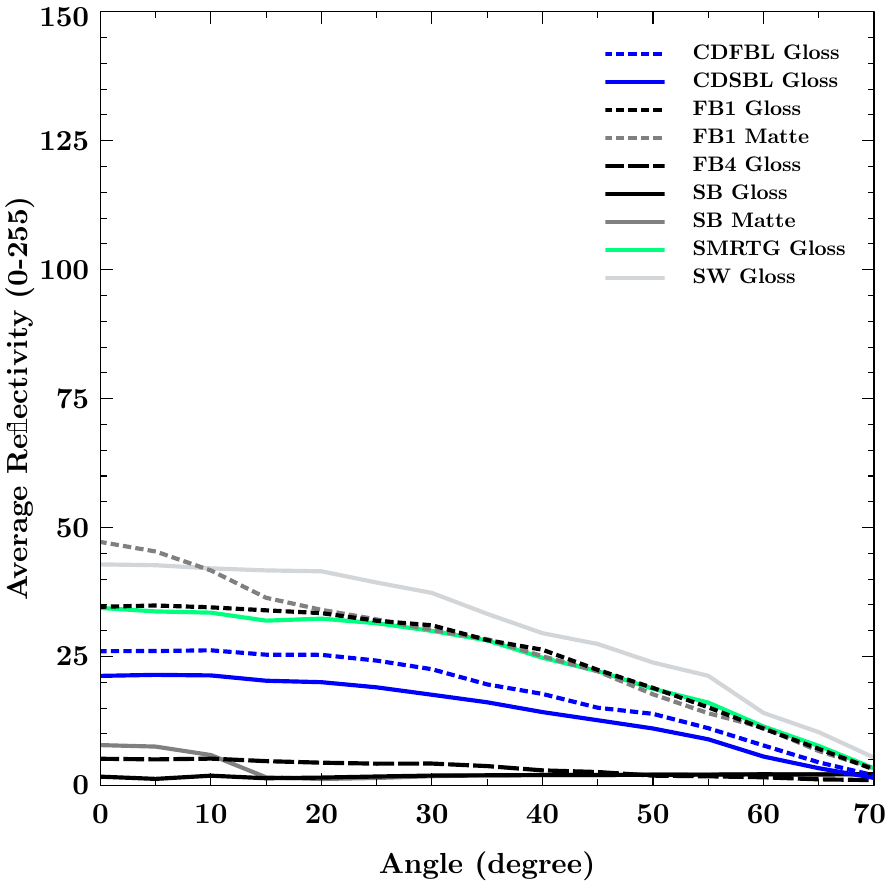}}
	\hspace{1pt}
	\subfloat[]{\label{fig:metallic_16} \includegraphics[width=0.48\columnwidth]{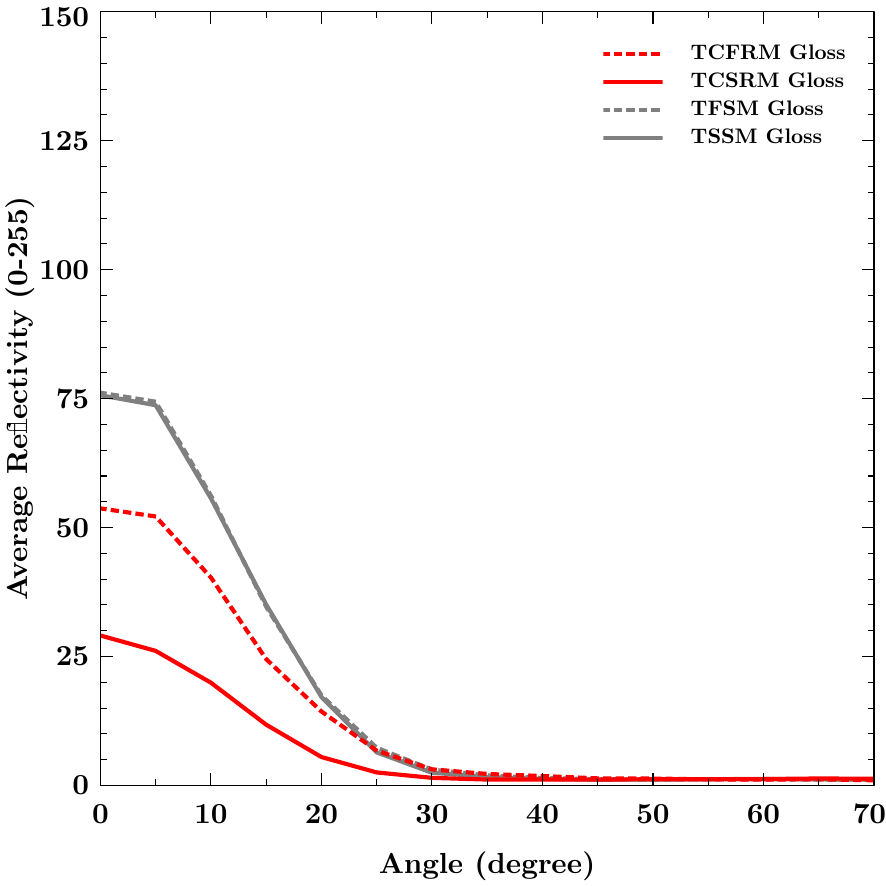}}
    \caption{VLP-16 Average reflected intensity vs. Angle of incidence at 2.5m \textbf{(a)} Non-metallic   \textbf{(b)} Metallic paints}
    \label{fig:VLP_16_2.5m}
\end{figure}

The tests using the VLP-16 shows similar trends to the ones seen for the VLS-128 at the selected distance of 2.5m. This is depicted in Fig. \ref{fig:VLP_16_2.5m}, where the two distinct groups of paints can also be observed in the data captured at a distance of 2.5m. However, one other observation we make is that the maximum intensity for all paint panels are lower for the VLP-16 as compared to the VLS-128. This is expected as the VLP-16 is a lower power LiDAR sensor with a smaller advertised range. It is meant for close-range applications. Hence, the emitted and consequently, the reflected laser beams would be of lower intensities than those from the VLS-128. 

\subsubsection{Tests using Leddartech Pixell}

\begin{figure}[h!]
    \centering
	\captionsetup{justification=centering}
	\subfloat[]{\label{fig:non-metallic_16} \includegraphics[width=0.48\columnwidth]{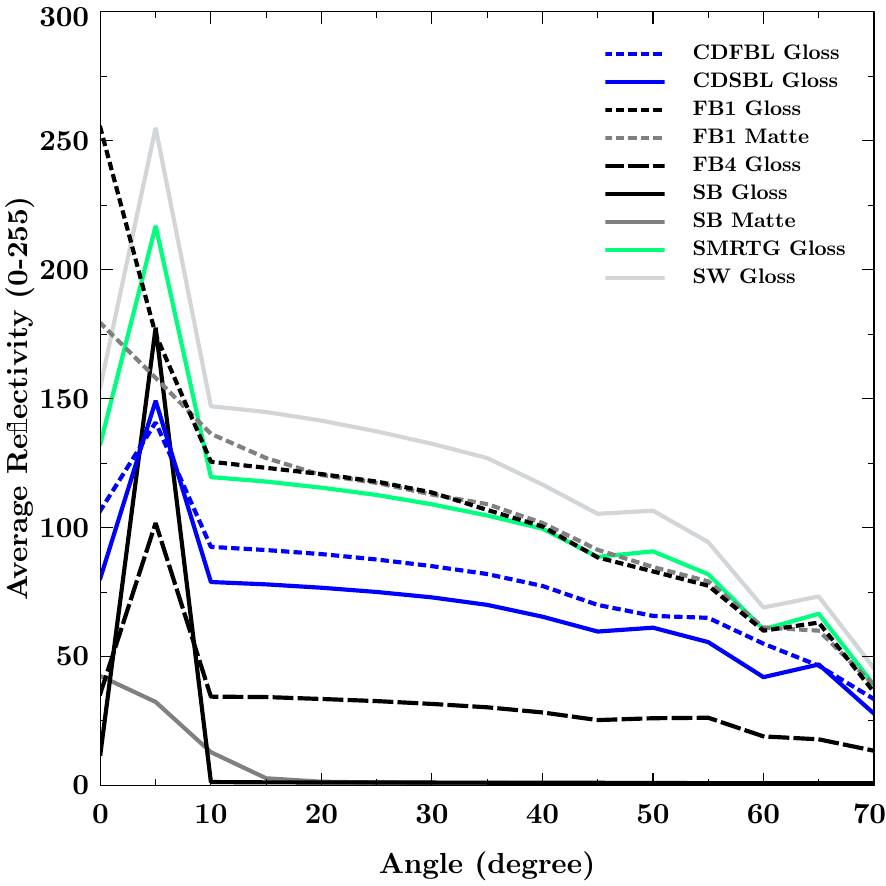}}
	\hspace{1pt}
	\subfloat[]{\label{fig:metallic_16} \includegraphics[width=0.48\columnwidth]{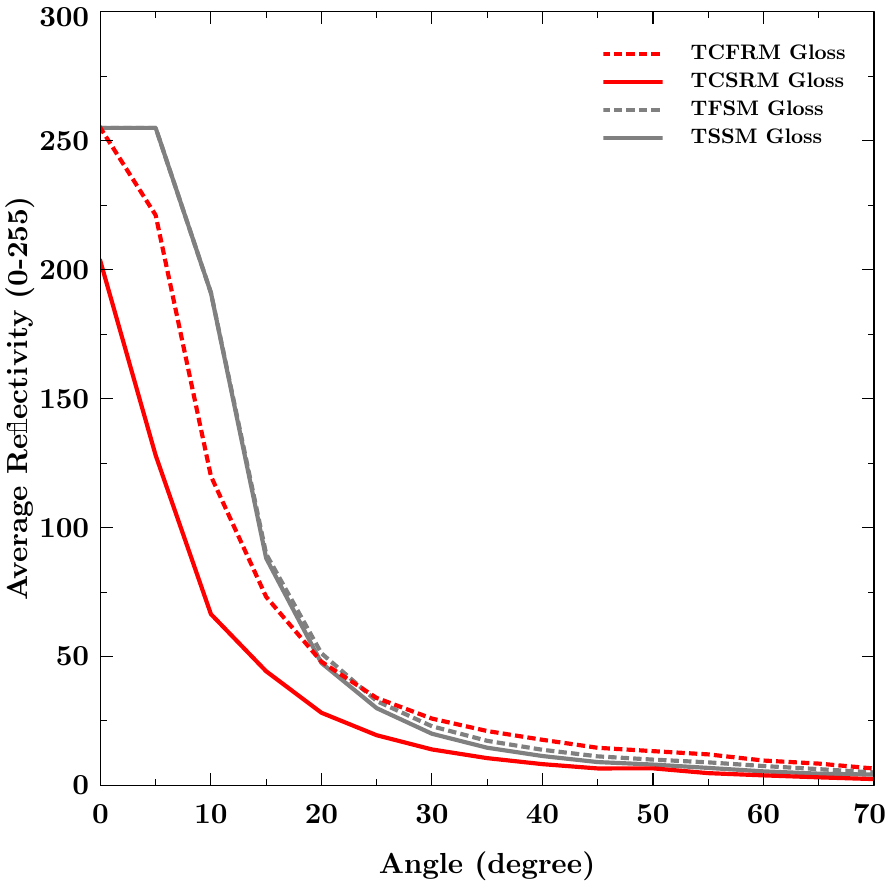}}
    \caption{Leddartech Pixell Average reflected intensity vs. Angle of incidence at 2.5m \textbf{(a)} Non-metallic   \textbf{(b)} Metallic paints}
    \label{fig:Pixell_2.5m}
\end{figure}

In order to provide some variation on the type of LiDAR sensor being used, we also tested the Leddartech Pixell, which is a solid state sensor. The results of these tests for the distance of 2.5m is shown in Fig.~\ref{fig:Pixell_2.5m}. The intensity range of the Leddartech Pixell is 0 -- 262143. The intensity range is linear until the detected intensity reaches a saturation point of 2048. From 2048, the received signal is considered to be `saturated' upon which the sensor's signal processing algorithm attenuates the received signal, and outputs a non-linear value which may be up to 262143. The graph in Fig.~\ref{fig:Pixell_2.5m} is constructed from data taken from the linear range of 0 -- 2048 and normalised to a range of 0 -- 255 to give an accurate comparison with the other LiDAR sensor outputs. After normalisation, the trend seen in Fig.~\ref{fig:Pixell_2.5m} is generally similar to the trends for both Velodyne LiDARs, where two distinct groups of paints are being observed. However, one difference we notice is the sudden spike in intensity for non-metallic paints at 5\degree. We suspect that this might be due to specular reflection of light on the panels being picked up by the sensor at this angle, whereas this is recorded at 0\degree ~for both Velodyne LiDARs.

\begin{figure}[h!]
	\centering
	\captionsetup{justification=centering}
	\subfloat[]{\label{fig:vertical_FOV} \includegraphics[width=0.61\columnwidth]{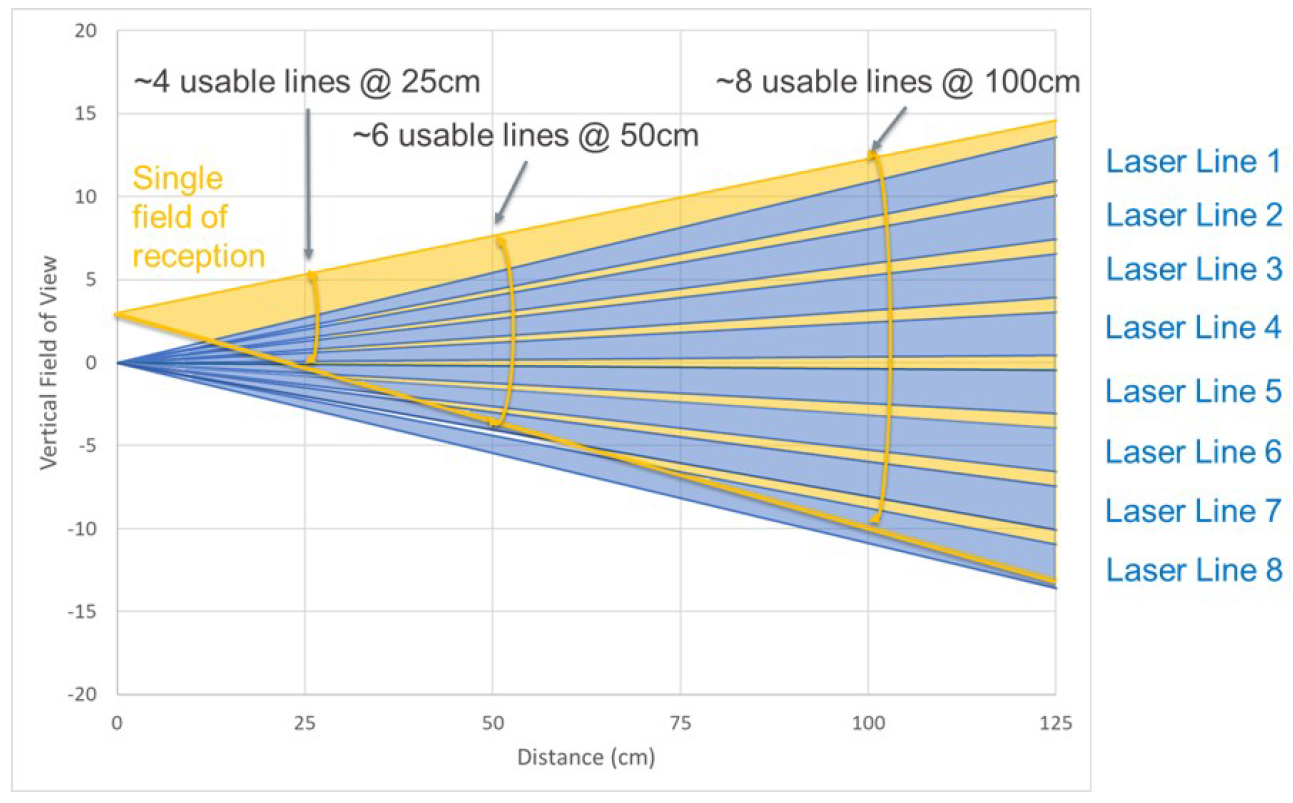}}
	\hspace{1pt}
	\subfloat[]{\label{fig:2_zones} \includegraphics[width=0.365\columnwidth]{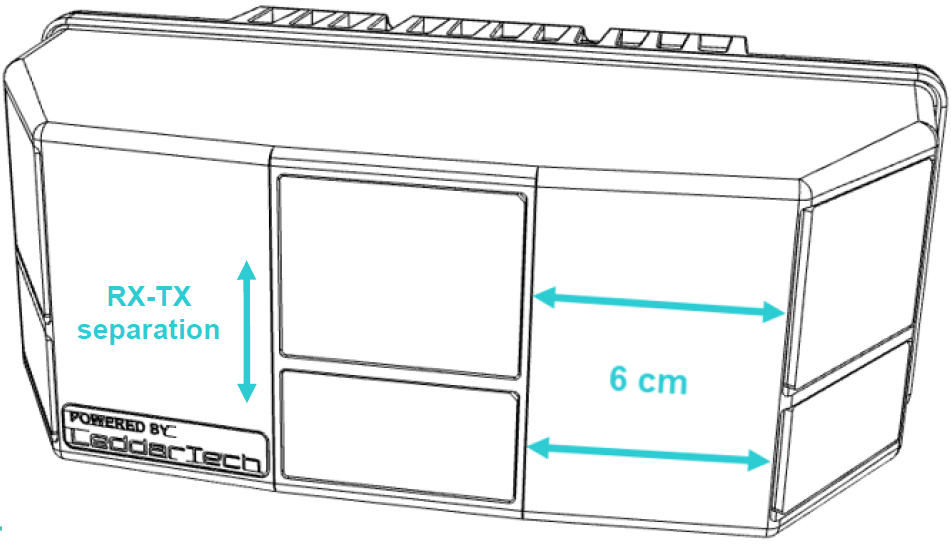}}
	\caption{\textbf{(a)} Pixell vertical emission coverage   \textbf{(b)} Pixell distance between two zones \cite{pixell_guide}}
	\label{fig:pixell_config}
\end{figure}

One possible explanation for this phenomenon could be because of the Leddartech Pixell's bi-static optical design. The LiDAR sensor has a physical separation between the transmitter window and receiver window, illustrated in Fig. \ref{fig:2_zones}, which creates a variable vertical overlap at close range, as shown in Fig. \ref{fig:vertical_FOV}. The receiver has a single wide Field-of-View (FoV) depicted in orange in Fig. \ref{fig:vertical_FOV}, while the emitting section consists of 8 laser lines covering the entire horizontal FoV. However, each line scans a portion of the vertical FoV and at a very close range, only the top-emitting segments overlap with the receiving FoV \cite{Pixell_guide}. This particular design, together with the lower resolution of the Leddartech Pixell as compared to both Velodyne LiDARs could be the reason for a difference in data observed.

\clearpage

\subsection{Analysis}

Based on the results recorded for all three LiDAR sensors, it is clear that LiDAR performance is affected by the type of paint used as the test target. The type of paints tested can be split into the following groups: LiDAR functionalised paints vs standard paints, metallic paints vs non-metallic paints and glossy vs matte paints. These distinct separations of paint types and their properties can be observed after analysis of our data. LiDAR functionalised paints are definitely more visible to LiDAR sensors as compared to standard paints since they almost always return higher intensity values across all three LiDAR sensors.

On the difference in intensity between light and dark colours, we focus on Standard White paint (SW\_Gloss) and Standard Black paint (SB\_Gloss) as an example. The SW\_Gloss paint has consistently high intensity readings among all paint panels across all three LiDAR sensors. Whereas, the SB\_Gloss paint has the lowest intensity readings among all paint panels across all three LiDAR sensors.

Looking at metallic paints as compared to non-metallic ones, as mentioned in previous sections, there are distinct intensity vs. angle trends. After consultation with our paint partner, this trend is caused by the addition of metallic pigments which are present in metallic paints. These are thin platelet-shaped aluminium particles which act like small mirrors and cause direct light reflections causing a sudden drop in reflected intensity when changing the incident angle. This causes the sharp drop in reflected intensity for metallic paints when angle of the panel increases.

\begin{figure}[htb]
    \centering
    \centerline{
        \includegraphics[scale=0.6]{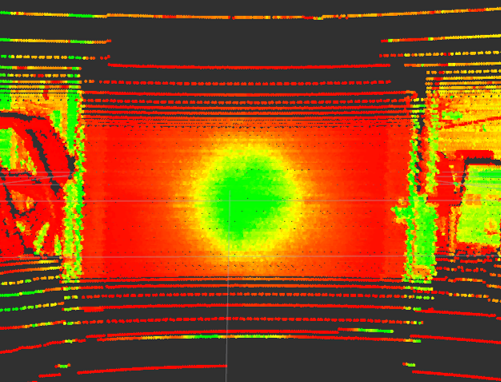}
    }
    \caption{Specular reflection observed for SB\_Matt panel at 0\degree}
    \label{fig:sb_matte}
\end{figure}

Lastly, matte paints tend to show higher intensity initially at 0 -- 15\degree ~as compared to their glossy counterparts. This seemed to be due to the nature of the specular reflection of the matte paint panels, at angles close to zero (normal to the incident LiDAR beams). This was observed during our tests when the panel is normal to the LiDAR beams, a large circular area of higher intensity was found at the center of a matte paint panel as shown in Fig. \ref{fig:sb_matte}, compared to their glossy counterparts. This area of higher intensity disappeared when the paint panel was shifted to a steeper angle.

\clearpage

\section{Conclusion}

This white paper provides an overview of the impact of automotive paint on LiDAR performance and presents the methodology used to test LiDAR performance against different automotive paint coatings. From the results and analysis of the tests, we have gained some insight on which types of paint panels have enhanced LiDAR visibility. Our results corroborate what was previously, namely, that white paints are more visible to a LiDAR sensor than black paints.

In addition, we have a better understanding of the effects of metallic vs non-metallic paints on LiDAR performance. Metallic paints tend to produce large intensity changes depending on the angle of orientation of the panel. On the other hand, non-metallic paints exhibit gradual intensity changes when the paint panel angle of orientation is adjusted.

The Velodyne VLS-128 was tested at selected distances to the target from 2.5m up to 30.0m, in order to gain an understanding on the effects of target distance on LiDAR performance. We observed that the trends are very similar for each panel, with just a slight drop in intensity values as the distance increases. This slight drop in intensity is likely due to the loss of energy in the laser beams as it travels through the atmosphere.

Verification testing was done to confirm that LiDAR sensor performance when rotating the panel with respect to the elevation angle is similar to when it is rotated with respect to the azimuth angle. Verification testing was
also done to confirm that testing indoors in artificial lighting is representative of LiDAR results outdoors under sunlight. Whilst slightly higher absolute values where observed indoors than in outdoor testing, the trends of intensity with change in angle were very similar.

Finally, we also see the potential of LiDAR functionalised paint types being applied onto road vehicles. Looking at our results, LiDAR functionalised paint types consistently show higher LiDAR intensity readings than their standard paint counterparts. The addition of LiDAR functional pigments to paint improves the ability of a LiDAR sensor to return a high intensity signal and enhances LiDAR performance. Therefore, it is worth exploring the possibility of using LiDAR functionalised paints on road vehicles in the future when the number of AVs on the road will undoubtedly increase. However, perception algorithms should continue to be robust enough by being able to detect and classify vehicles with conventional paints and to not only detect higher intensity reflections as this would limit an AV's perception performance when standard paints are used as compared to LiDAR functionalised paints.

\section{Acknowledgements}
\label{sec:Acknowledgements}

We acknowledge our current partners who are supporting us in this project. In particular, we thank Nagase \& Co., Ltd., for providing us paint samples that we have made use of for our initial experimentation. We also thank NIPSEA Technologies Pte. Ltd. for providing us with a diverse set of automotive paint panels and the ongoing collaboration. Furthermore, we thank LeddarTech Inc., for loaning the LeddarTech Pixell LiDAR, providing technical consultancy and ideas regarding testing and verification of commercial LiDARs.

This research/project is supported by the National Research Foundation, Singapore, and Land Transport Authority under Urban Mobility Grand Challenge (UMGC-L010). Any opinions, findings and conclusions or recommendations expressed in this material are those of the author(s) and do not reflect the views of National Research Foundation, Singapore, and Land Transport Authority.

\cleardoublepage

\printbibliography 

\clearpage
\renewcommand{\thefigure}{\thesection.\arabic{figure}}  
\setcounter{figure}{0} 
\renewcommand{\thetable}{\thesection.\arabic{table}}    
\setcounter{table}{0} 

\end{document}